\documentclass[11pt]{article}

\usepackage[utf8]{inputenc}
\usepackage[T1]{fontenc}
\usepackage{lmodern}
\usepackage{microtype}

\usepackage[a4paper, top=2.5cm, bottom=2.5cm, left=2.8cm, right=2.8cm]{geometry}

\usepackage{amsmath}
\usepackage{amssymb}

\usepackage{booktabs}
\usepackage{longtable}
\usepackage{array}
\usepackage{multirow}
\usepackage{calc}          % for \real{} in longtable column specs

\usepackage{graphicx}
\usepackage{xcolor}

\usepackage[
  colorlinks=true,
  hyperfootnotes=false,
  linkcolor=blue!70!black,
  citecolor=blue!70!black,
  urlcolor=blue!70!black,
  pdfauthor={Bo Zou, Chao Xu},
  pdftitle={BC Protocol: Structured Dual-Expert Dialogue for Eliciting
            High-Quality Chain-of-Thought Post-Training Data},
]{hyperref}
\usepackage{bookmark}

\usepackage{parskip}       % space between paragraphs instead of indent
\usepackage{setspace}
\title{\textbf{BC Protocol: Structured Dual-Expert Dialogue for Eliciting\\
High-Quality Chain-of-Thought Post-Training Data}}
\author{Bo Zou \and Chao Xu}
\date{2026}

\newcounter{none}
\providecommand{\tightlist}{\setlength{\itemsep}{0pt}\setlength{\parskip}{0pt}}
\begin{document}
\maketitle

\begin{abstract}

High-quality expert chain-of-thought (CoT) data is one of the core bottlenecks in large language model (LLM) post-training. Existing data production methods each have structural limitations: crowdsourced annotation lacks deep reasoning paths; expert solo writing is constrained by the ``expert blind spot''---experts structurally skip reasoning steps they consider obvious; RLHF only produces preference signals rather than reasoning chains.

This paper proposes the BC Protocol---a structured dual-expert elicitation method for LLM post-training data production. The method carefully pairs a domain expert (crystallized intelligence) with a knowledge engineer (fluid intelligence), systematically externalizing the expert's implicit judgments as natural language reasoning chains. We introduce the Participant Aptitude Model, which defines six participant characteristic dimensions that affect elicitation quality. ``Calibrated Ignorance'' is an original concept proposed in this paper. We further propose ``Selection-over-Prescription'' as a methodological principle: for implicit knowledge elicitation tasks, investing quality-control resources in personnel selection yields a higher return than investing the same resources in process design.

In a controlled experiment in the narrative fiction domain, we directly compared CoT produced by BC Protocol dual dialogue (Group A, \(n=20\)) against CoT written independently by the same domain expert (Group B, \(n=20\)). Three cross-vendor judge models---GPT-4o, Claude Opus 4.5, and Gemini 2.5 Pro---conducted blind evaluation across five dimensions (600 ratings total). Results show that the BC Protocol achieves an overwhelming advantage in ``naturalness of reasoning process'' (Group A mean 4.80 vs.~Group B mean 1.30, \(p=2.4\times10^{-8}\), Cliff's \(\delta=1.0\)). It shows trending advantages in ``counterfactual density'' and ``reasoning chain completeness,'' though these do not reach \(p<0.05\) significance at the current sample size. Group B shows a significantly higher ``information density''---a contrast that confirms our core judgment: Expert Solo outputs are post-hoc conclusions compressed into high-density terminology, while BC dialogue outputs are live reasoning that preserves trial-and-error nodes. One hour of voice dialogue can produce 10--20 CoT samples directly usable for post-training. This provides a replicable, high-quality data production pipeline for vertical-domain LLM alignment at controllable engineering cost. We further argue that dual dialogue driven by epistemic vigilance naturally accommodates ``counterfactual probing''---the highest-intensity elicitation action. As a result, the BC Protocol synchronously produces high counterfactual density reasoning chains at the collection stage. Compared to the current industry practice of post-hoc counterfactual data augmentation (4--5 minutes per manual counterfactual rewrite, Kaushik et al., 2020), this synchronous elicitation path has a significant marginal cost advantage.

\end{abstract}

\section{Introduction}\label{introduction}

\subsection{The Data Bottleneck in LLM Alignment}\label{the-data-bottleneck-in-llm-alignment}

The key bottleneck in LLM capability improvement is shifting from model architecture and parameter scale toward data quality. Zhou et al.~(2023)'s LIMA experiment revealed this trend in a striking way: fine-tuning LLaMA with only 1,000 carefully selected high-quality samples achieved performance comparable to models trained on tens of thousands of crowdsourced samples. This finding means that, in fine-tuning, the leverage ratio between data ``quality'' and ``quantity'' is extremely lopsided. A small number of genuinely high-quality samples releases far greater alignment power than large-scale but inconsistent data.

Yet ``high quality'' remains a poorly defined black box in the current research literature. Most work equates it with ``well-formatted, diverse instructions, fluent responses.'' Few papers ask a more fundamental question: Who produced this data, under what cognitive conditions, and in what way? Is the data production process itself the overlooked core variable that determines final quality?

\subsection{Structural Limitations of Existing Data Production Methods}\label{structural-limitations-of-existing-data-production-methods}

Examining this question exposes structural weaknesses in the four dominant data production approaches---crowdsourced annotation, expert solo writing, RLHF, and synthetic data. We analyze these structural limitations individually in §2.1. Here we only note one shared blind spot: all four approaches optimize data quantity, format, and diversity while largely ignoring the \textbf{design of the production process itself}---that is, the cognitive conditions and interaction structure under which expert knowledge is elicited.

\subsection{Our Proposal}\label{our-proposal}

This paper proposes the BC Protocol\footnote{The name ``BC'' comes from the two founding practitioners of this protocol. B (Bo) represents the knowledge engineer/elicitor with high fluid intelligence. C (Chao) represents the domain expert with high crystallized intelligence and deep domain experience.}---a structured dual-dialogue elicitation method. It carefully pairs a domain expert with a knowledge engineer who have complementary cognitive profiles, systematically externalizing the expert's implicit judgments as natural language chain-of-thought (CoT) reasoning chains.

The key insight of the BC Protocol is this: the quality of CoT data depends not only on ``how strong the expert is,'' but also on \textbf{the design of the elicitation process itself}. To address this, we propose the \textbf{Participant Aptitude Model}, which defines six participant characteristic dimensions that affect elicitation quality. The model shifts the center of quality assurance from ``standardizing the operating procedure'' to ``rigorously selecting the participants.'' In terms of engineering feasibility, the BC Protocol uses voice dialogue---the expert's most natural mode of output---to produce high-density reasoning chains at controllable cost. One hour of dialogue can produce 10--20 CoT samples directly usable for post-training. The output is natural language that LLMs can use natively, with no format conversion needed.

From a data engineering perspective, the BC Protocol forms a Standardized Data Generation Pipeline: the input is a B--C pair screened by the Participant Aptitude Model; the middle stage is a voice dialogue process governed by a semi-structured protocol; the output is structured CoT samples directly usable for LLM post-training. Each stage of this pipeline---personnel selection criteria (§3.2), the dialogue protocol (§3.3--§3.4), and post-processing specifications (§3.5)---has been explicitly defined. This makes the pipeline replicable across domains and deployable at scale.

\subsection{Contributions}\label{contributions}

The main contributions of this paper are as follows:

\begin{enumerate}
\def\labelenumi{\arabic{enumi}.}
\item
  We propose the BC Protocol methodology---a structured dual-expert elicitation framework for LLM post-training data production.
\item
  We propose the Participant Aptitude Model, identifying and defining six participant characteristic dimensions that affect knowledge elicitation quality. ``Calibrated Ignorance'' is an original concept introduced in this paper.
\item
  We propose ``Selection-over-Prescription'' as a methodological principle. We argue that for high-order tasks like implicit knowledge elicitation---which depend fundamentally on the elicitor's implicit capabilities---investing quality-control resources in personnel selection yields a higher return on investment than investing them in designing more refined operating procedures.
\item
  We conduct a controlled experiment in the narrative fiction domain (Group A: BC dialogue vs.~Group B: Expert Solo, \(n=20\) each; 3 cross-vendor LLM judge models; 5 dimensions; 600 blind ratings total). Results show an overwhelming advantage in ``naturalness of reasoning process'' (Cliff's \(\delta=1.0\), \(p=2.4\times10^{-8}\)) and consistent trending advantages in ``counterfactual density'' and ``reasoning chain completeness''; we also honestly report Expert Solo's reverse advantage in ``information density'' and explain this contrast in §5.2. We further position the two authors as an \textbf{untrained BC baseline} (i.e., without targeted training in high-order epistemic actions such as counterfactual probing and meta-premise clarification). Accordingly, we propose a two-level framework distinguishing ``structural gains'' (significant without training---D3) from ``trainable gains'' (requiring targeted epistemic training---D1/D2/D5; see §5.2.2, §5.3, §5.4).
\item
  We integrate the methodology into an end-to-end Standardized Data Generation Pipeline. Its core stages---participant selection, dialogue execution, and post-processing---each have explicit input-output specifications and quality control nodes, providing an engineering foundation for cross-domain replication and scaled deployment.
\item
  We propose \textbf{counterfactual density} as an observable quality indicator for high-quality CoT data. We argue that the counterfactual probing naturally driven by epistemic vigilance in the BC Protocol constitutes a low-cost path that produces high counterfactual density at the collection stage---relative to post-hoc counterfactual data augmentation (Kaushik et al., 2020)---without incurring a separate offline counterfactual rewriting cost for the same data.
\end{enumerate}

\section{Related Work}\label{related-work}

This section reviews three research lineages directly related to the BC Protocol: data engineering practices in LLM post-training (§2.1), knowledge elicitation methodology (§2.2), and the theoretical foundations of tacit knowledge and expert cognition (§2.3). We then clarify the relationship between this paper and existing work in §2.4.

\subsection{Data Engineering in LLM Post-Training}\label{data-engineering-in-llm-post-training}

Recent years have seen the key bottleneck in LLM capability improvement shift from model architecture and parameter scale toward training data quality. Zhou et al.~(2023)`s LIMA experiment demonstrated that fine-tuning LLaMA with only 1,000 carefully selected high-quality samples achieves performance comparable to---or better than---models trained on tens of thousands of crowdsourced samples. This finding has drawn researchers' attention to a deeper question: how should high-quality post-training data actually be produced?

Existing data production methods fall roughly into four categories, each with structural limitations in eliciting deep reasoning chains. \textbf{Crowdsourced annotation} (exemplified by the Scale AI model) is low-cost and scalable, but annotators are not domain experts. They can only produce shallow preference judgments without reasoning paths. \textbf{Expert solo writing} is constrained by the ``expert blind spot''---experts skip intermediate reasoning steps they consider ``obvious'' and produce polished conclusions rather than the derivation process. \textbf{RLHF} (Ouyang et al., 2022) provides preference signals through ranking. It teaches models ``what to do'' but not ``how to think.'' The output is an A\textgreater B ranking without a reasoning chain explaining why A is better than B. \textbf{Synthetic data} (e.g., Self-Instruct, Wang et al., 2023) is essentially cyclic distillation of models teaching models, unable to inject new expert cognition.

The shared blind spot across all four approaches is: \textbf{excessive focus on the ``static result attributes'' of data, while ignoring the ``dynamic cognitive process.''} Researchers work hard to optimize data quantity, format, and diversity. Yet they rarely ask: how was the professional logic behind this data generated? What reasoning path and decision trade-offs did the annotator experience while producing the answer? The BC Protocol enters from precisely this neglected dimension.

\subsection{Failure Mode Analysis of Prevailing Annotation Pipelines}\label{failure-mode-analysis-of-prevailing-annotation-pipelines}

§2.1 outlined the structural limitations of four dominant post-training data production approaches. Remaining at the level of ``approaches,'' however, is insufficient to fully situate the specific pain points that the BC Protocol's design addresses. This section provides a structured analysis of the observable failure modes exhibited by today's dominant ``crowdsource + rubric-driven'' annotation pipeline---typified by large-scale outsourced writing-type CoT annotation. Our focus on these failure modes is not industrial criticism; rather: \textbf{the legitimacy of any methodology is partly built on a clear diagnosis of the failure mechanisms of current alternative paths.}

We organize these failure modes into five mutually independent but mutually reinforcing mechanisms. Each corresponds to a specific design choice in the BC Protocol (see §3 for the corresponding responses):

{\def\LTcaptype{none} % do not increment counter
\begin{longtable}[]{@{}
  >{\raggedright\arraybackslash}p{(\linewidth - 4\tabcolsep) * \real{0.3333}}
  >{\raggedright\arraybackslash}p{(\linewidth - 4\tabcolsep) * \real{0.3333}}
  >{\raggedright\arraybackslash}p{(\linewidth - 4\tabcolsep) * \real{0.3333}}@{}}
\toprule\noalign{}
\begin{minipage}[b]{\linewidth}\raggedright
Failure Mechanism in Current Pipelines
\end{minipage} & \begin{minipage}[b]{\linewidth}\raggedright
Observable Phenomena (drawn from public industry interviews and practitioner accounts)
\end{minipage} & \begin{minipage}[b]{\linewidth}\raggedright
Corresponding BC Protocol Design Choice
\end{minipage} \\
\midrule\noalign{}
\endhead
\bottomrule\noalign{}
\endlastfoot
\textbf{M1. Rule-makers disconnected from domain judgment} & Project managers who design annotation rules are often temporarily seconded across departments and lack judgment experience in the target domain (e.g., narrative fiction). The resulting rules exhibit ``broad but operationally low-resolution'' characteristics, typically assuming a ``correct answer'' exists for open-ended contested questions (e.g., ``the true motivation behind a plot turn''). & Selection-over-Prescription: C, who possesses real-world domain judgment, leads the reasoning; B serves only as an epistemic vigilance role. Rules are not preset---they emerge during dialogue and are named by C using native concepts (§3.2, §3.3). \\
\textbf{M2. Forced application of preset label libraries causes semantic compression distortion} & Character traits are forced into 20--30 predefined labels (e.g., ``scheming,'' ``doormat,'' ``manipulative''), unable to capture real character arcs and variants in actual creative work. Annotators are forced to compromise between ``inaccurate but compliant'' and ``accurate but out-of-bounds,'' systematically degrading high-order bits. & No preset label libraries: CoT allows judgments to emerge naturally during dialogue and preserves C's native conceptual naming. The post-processing stage (§3.5) applies structuring but not categorization. \\
\textbf{M3. Hard token/word count caps inversely compress semantics to distortion} & Hard length constraints set to align with model context limits (e.g., ``a 1,000-character chapter outline must not exceed 350 characters'') force annotators to merge character pronouns, compress multi-character actions into single long sentences, and drop plot causal chains. The word count constraint inversely reconstructs the semantics themselves. & Recording → transcription → complete natural language reasoning is retained; reasoning chains are not length-truncated. §3.4 argues for the advantage of the voice medium over text in preserving reasoning completeness. \\
\textbf{M4. ``Standard answer'' illusion and flattened emotion labels} & Paragraph function and emotional quality are forced into predefined categories (e.g., nine paragraph functions, a fixed set of emotion words). Complex, multi-layered, self-contradictory real emotions are compressed into flat labels like ``sad'' or ``angry.'' In the open-ended domain of creative judgment, this compression is itself a distortion. & Explicitly acknowledges that creative judgment has no unique correct answer (§3.3). CoT records C's judgment reasoning and its premises, not a ``right-or-wrong binary.'' \\
\textbf{M5. Unidirectional scoring incentive misalignment and cognitive depletion at the pipeline's end} & Annotators operate in a ``submit → quality-rejected → redo'' unidirectional feedback structure. Incentive misalignment leads to observable cognitive depletion within months: inability to detect ``AI-feeling,'' reverse degradation of writing ability. & Symmetric dual-expert dialogue: no scoring relationship between B and C; epistemic challenges are bidirectional. The BC interaction structure is designed to be cognitively beneficial to both parties (§3.2.3, §3.3). \\
\end{longtable}
}

It should be noted that the failure modes identified here do not constitute criticism of any specific company or team. These mechanisms are \textbf{structural consequences} of the ``crowdsource + rubric-driven'' pipeline in high-tacit-knowledge domains such as creative judgment---not isolated problems caused by individual management choices. The existence of these failure modes shows precisely that: in creative judgment, the data pipeline bottleneck lies not in ``finding more cheap annotators'' or ``writing more refined annotation rules,'' but in \textbf{finding the few experts with deep domain judgment and designing an elicitation structure that faithfully externalizes their implicit judgments}. This is exactly the methodological response provided by the BC Protocol in §3.

\subsection{Knowledge Elicitation Methodology}\label{knowledge-elicitation-methodology}

The systematic extraction of knowledge from expert minds is not a new problem. During the expert-systems era of the 1980s--1990s, the field of knowledge engineering developed a range of classical methods. \textbf{Structured interviews} guided experts to output decision logic step by step through preset question frameworks. \textbf{Think-Aloud Protocol} (Ericsson \& Simon, 1993) required experts to verbalize their thought processes in real time while executing a task. \textbf{Critical Decision Method} (Klein et al., 1989) focused on expert judgment in high-pressure situations, reconstructing reasoning trajectories through retrospective interviews.

These methods were designed to construct IF-THEN production rules to drive the inference engines of expert systems. That era defined ``knowledge'' as something that could be encoded as formal rules. However, LLM post-training needs not formal rules but \textbf{natural language reasoning chains}. This fundamentally changes the quality definition of the ``elicitation product.'' Traditional methods pursued the precision and completeness of rules. What we pursue is the naturalness, completeness, and information density of the reasoning process.

\subsection{Tacit Knowledge and Expert Cognition}\label{tacit-knowledge-and-expert-cognition}

Polanyi (1966) proposed a challenge that remains unresolved: ``We know more than we can tell.'' Much human professional expertise exists as tacit knowledge, which its holder cannot fully verbalize through active introspection. Nathan and Petrosino (2003) further revealed the cognitive mechanism---the ``Expert Blind Spot'': as domain experience accumulates, many decision processes that once required step-by-step reasoning become compressed into automated pattern recognition. The expert ``knows the answer at a glance,'' but when asked to explain, all they can output is the final conclusion. The many micro-reasoning steps in between have sunk below the threshold of consciousness.

These two concepts directly support the core assumption of the BC Protocol: the fact that an expert ``cannot say it'' does not mean the knowledge does not exist---it means that \textbf{specific external conditions} are needed to activate and externalize it. Simply ``asking'' an expert is not enough. What is needed is a way to break through their automated pattern recognition and force experts to ``unpack'' their intuitive judgments backwards into explicit reasoning chains. This is the fundamental reason the BC Protocol exists.

\subsection{Relationship to Existing Work}\label{relationship-to-existing-work}

This paper stands at the intersection of knowledge engineering and LLM data engineering, connecting the core problem awareness of 1980s knowledge elicitation methodology with the natural language post-training requirements of the 2020s. Compared to existing work, this paper's contributions operate at three levels.

\textbf{First, from a ``checklist of elicitor qualifications'' to a ``causal mechanism model.''} To be clear, 1980s knowledge engineering literature did not entirely ignore the question of elicitor qualifications. Classic textbooks extensively discussed the interpersonal skills, cognitive styles, and communicative traits that knowledge engineers should possess (Waterman, 1986; McGraw \& Harbison-Briggs, 1989), explicitly defining the knowledge engineer as a ``research tool'' and acknowledging that elicitation product quality is filtered through the elicitor's personal qualities. A few pioneering empirical studies even directly tested the effect of elicitor traits on elicitation quality: Mykytyn et al.~(1994) constructed a quantitative scale of 26 behavioral skills and extracted five core competency factors. However, such work represented an extremely small fraction of total 1980s--1990s knowledge engineering research. The vast majority of empirical attention was drawn to ``method-centric'' paradigms---comparing the relative efficiency of different acquisition techniques (e.g., structured interview vs.~repertory grid vs.~protocol analysis) (Hoffman, 1987; Cooke, 1994). In these studies, elicitor personal traits were typically treated as ``interference noise'' to be controlled for through standardized procedures, not as the core independent variable for exploring causal mechanisms of elicitation quality. The BC Protocol's Participant Aptitude Model (§3.2) inherits the concerns of these pioneering studies but achieves deeper development in the following ways. First, it integrates scattered theoretical checklists and local experiments into a systematic, internally coherent evaluative framework. Second, it explicitly distinguishes hard constraints (whose absence is irreparable) from soft constraints (whose absence degrades efficiency). Third, it establishes a causal transmission path from cognitive mechanism to data quality for each dimension.

\textbf{Second, from the elicitor's ``tool attribute'' to ``the structure of interpersonal interaction.''} Classic literature's discussion of knowledge engineer qualifications ultimately cared about ``how to more efficiently extract rules from the expert's mind.'' The knowledge engineer was treated as a high-performance cognitive tool, and qualification requirements were essentially specifications for tool performance. The BC Protocol's Participant Aptitude Model enacts a paradigm shift: it moves the focus from ``the unilateral tool attribute of the elicitor'' to ``the dynamic quality of the two-person interaction structure.'' For example, Dimension 1 (truth-seeking epistemic orientation) is not a one-way skill requirement on either B or C alone---it is a constraint on the interaction relationship itself. One party deviating from truth-seeking will infect the other through a chain reaction (§3.2.3). Dimension 6 (complementary information processing preferences) concerns how the cognitive tension between B and C generates synergistic output that exceeds the sum of the two individuals. This shift in focus from ``individual tool performance'' to ``two-person interaction structure'' is the fundamental divide between this paper and classical knowledge engineering literature.

\textbf{Third, redefining the output product.} The goal product of traditional knowledge engineering was formal IF-THEN rules. The goal product of the BC Protocol is natural language chain-of-thought (CoT)---directly usable for LLM post-training with no format conversion needed. This fundamentally solves the incompatibility problem between traditional knowledge engineering output and LLM training formats.

\section{The BC Protocol Framework}\label{the-bc-protocol-framework}

\begin{quote}
\textbf{Core Claim.} For high-order tasks like implicit knowledge elicitation---which depend fundamentally on the elicitor's implicit capabilities---investing a larger share of resources in finding and selecting the right people yields a far higher return on investment than investing the same resources in designing more refined procedures. This ``\textbf{Selection-over-Prescription}'' methodological stance is the fundamental divide between the BC Protocol and traditional knowledge engineering methodology (§3.3 further elaborates this principle).
\end{quote}

This section presents the complete BC Protocol methodology: the overall workflow (§3.1); the ``Participant Aptitude Model'' defining the cognitive and personality traits required for effective participants (§3.2); the semi-structured elicitation guide governing the dialogue process (§3.3); the theoretical rationale for choosing voice over text as the elicitation medium (§3.4); and the post-processing workflow for converting raw dialogue into structured CoT samples (§3.5).

\subsection{Protocol Overview}\label{protocol-overview}

The BC Protocol is a structured dual-dialogue elicitation method. Its goal is to transform the expert's implicit reasoning processes into natural language chain-of-thought (CoT). The protocol pairs a domain expert (hereafter \emph{C}) with a knowledge engineer (hereafter \emph{B}), and systematically externalizes implicit knowledge through a series of voice dialogues governed by procedural and interpersonal rules.

The process begins with a \textbf{preparation phase}: B and C agree on the dialogue scope. B does not need to rehearse specific questions in advance---only to familiarize themselves with the domain's knowledge background (or simply select a B who already has basic familiarity with the domain). Next comes the \textbf{dialogue phase}: both parties engage as fully equal co-explorers, conducting a deep exchange around one of C's expert decisions. Finally, the \textbf{post-processing phase} transcribes and refines the recorded dialogue, distilling it into high-quality CoT data samples directly usable for LLM post-training. The specific rules and operational details of each phase are elaborated later (especially §3.3 and §3.5).

From a data engineering perspective, the three phases of preparation, dialogue, and post-processing form an end-to-end Standardized Data Generation Pipeline. Each node of the pipeline has explicit quality control mechanisms: the output of the preparation phase is a qualified B--C pair confirmed through the Participant Aptitude Model (§3.2); the output of the dialogue phase is raw audio containing complete prosodic and tonal cues; the output of the post-processing phase is structured CoT samples conforming to post-training format specifications. This pipeline design ensures that the BC Protocol is not merely a one-time research method but a data production infrastructure replicable and deployable across domains and expert teams.

\subsection{Participant Aptitude Model}\label{participant-aptitude-model}

The ``Participant Aptitude Model'' is the core theoretical contribution of this paper. It identifies six participant characteristic dimensions that affect the quality of CoT data elicitation. Establishing the ``Participant Aptitude Model''---rather than an ``operational procedure manual''---as the core contribution is itself a methodological stance. Its internal logic can be reduced to an epistemological recursion: if tacit knowledge cannot be fully externalized through procedural means, then the ability to ``elicit tacit knowledge'' is itself a tacit skill---one that equally cannot be reduced to a standard operating procedure. Since ``how to elicit excellent reasoning chains'' cannot be encoded as a replicable workflow, we shift focus: discover those who can in fact elicit excellent reasoning chains, and characterize their common traits. In doing so, the BC Protocol completes a methodological \textbf{engineering shift}\footnote{In epistemological terms, this engineering shift is structurally isomorphic to AI's own paradigm shift from expert systems to neural networks. In the expert-system era, researchers tried to explicitly encode expert reasoning logic as IF-THEN rules (corresponding to ``designing procedures''). The neural network paradigm abandoned explicit rule encoding and instead sought high-quality learning carriers that could implicitly acquire the same process (corresponding to ``finding the right people''). The BC Protocol replicates this deeper epistemological shift in the task of knowledge elicitation.}: the production of excellent reasoning chains is entrusted to excellent practitioners rather than excellent procedures---quality control shifts from ``designing processes'' to ``identifying practitioners.'' The Participant Aptitude Model is the concrete product of this engineering shift: it does not prescribe what to do, but strictly constrains \textbf{who} can do it.

Before elaborating on each of the six dimensions, we must first establish a foundational distinction from existing literature: a fundamental redefinition of participant roles.

\subsubsection{Role Redefinition: From Expert Interviewee to Co-Explorer}\label{role-redefinition-from-expert-interviewee-to-co-explorer}

Traditional knowledge elicitation is prone to two typical ``power asymmetry'' traps. One is \textbf{instrumentalizing the expert} (as in early knowledge engineering), where the knowledge engineer (B) acts as the dominant questioner and the expert (C) becomes a passive information repository (B\textgreater C). The other is \textbf{elevating the expert to unquestioned authority}, where the expert (C) takes on the role of an authoritative instructor and the engineer (B) acts as a passive recipient (C\textgreater B). The BC Protocol aims to structurally avoid both forms of asymmetric interaction.

In the BC Protocol, the domain expert (C) is not only a knowledge source but also a \textbf{co-excavator} who is an equal of B. C does not passively receive questions or unidirectionally transmit knowledge---instead, C works together with B to reconstruct their own implicit reasoning process.

Establishing equal interaction is necessary to prevent participants from entering a ``role performance'' state. People adjust their behavior in response to different social identity interactions (Goffman, 1959). Once asymmetry appears in the dialogue---whether B is overly dominant or excessively deferential toward C---C may enter ``performance mode'': systematically outputting polished, mature conclusions rather than naturally reconstructing the raw derivation process that preceded those conclusions. This creates a ``lossy compression'' of reasoning information.

To reduce this effect, the BC Protocol requires minimizing psychosocial distance created by status during dialogue---avoiding excessive use of honorifics and other expressions that reinforce hierarchy, in order to suppress unequal interaction tendencies. Only when both parties maintain an equal collaborative relationship are high-information-density CoT data more likely to be fully elicited.

Based on this, before discussing the six dimensions, one core consensus must be established: these are emphatically not unilateral skill requirements on either B or C---they are absolute baseline constraints imposed on both parties simultaneously.

\subsubsection{Constraint Classification: Hard Constraints and Soft Constraints}\label{constraint-classification-hard-constraints-and-soft-constraints}

The six dimensions do not carry equal weight. Based on their impact on the success or failure of the protocol, we divide them into two constraint levels:

\textbf{Hard constraints} are dimensions whose absence directly causes the protocol to fail entirely. The resulting data is systemically contaminated and almost impossible to repair in post-processing. Two dimensions belong to this category: \textbf{Dimension 1 (truth-seeking epistemic orientation)}---if either party's deep subconscious slides toward ``impression management'' (manifesting as defensive justification or arrogant showmanship) rather than ``approaching the truth,'' the final output will be fatally contaminated by highly convincing ``rationalization noise'' or ``performative reasoning''; and \textbf{Dimension 2 (epistemic vigilance)}---if B cannot identify reasoning gaps in C's output, the mechanism that triggers targeted follow-up questioning is disabled, and the dialogue irreversibly degrades into an ordinary interview.

\textbf{Soft constraints} are dimensions that significantly improve elicitation quality but whose absence does not cause the protocol to collapse entirely. Under suboptimal conditions, the dialogue can still produce usable data---but with reduced efficiency and depth in externalizing information. The remaining four dimensions belong to this category: Dimension 3 (complementary intelligence profiles), Dimension 4 (calibrated cognitive gap), Dimension 5 (calibrated ignorance), and Dimension 6 (complementary information processing preferences).

The core practical implication of this classification is: hard constraints should be used for \textbf{one-veto participant screening} (failure to meet them results in immediate exclusion); soft constraints should be used for \textbf{pairing optimization} (the higher the fit between both parties, the better the signal-to-noise ratio of the jointly produced CoT data).

\subsubsection{Dimension 1: Truth-Seeking Epistemic Orientation (Hard Constraint; Absolute Baseline for Both B and C)}\label{dimension-1-truth-seeking-epistemic-orientation-hard-constraint-absolute-baseline-for-both-b-and-c}

This is the most critical and non-negotiable dimension in the Participant Aptitude Model. It is also the key characteristic distinguishing the BC Protocol from ordinary knowledge elicitation approaches. Without this dimension, elicitation quality degrades sharply or fails entirely.

\textbf{Concept definition.} ``Truth-seeking epistemic orientation'' means that participants' primary motivation in the dialogue must be ``exploring objective cognition,'' not ``protecting their own authority'' or ``winning the argument.'' Galef (2021) captures this as the difference between the ``scout mindset'' and the ``soldier mindset'': the former treats exposure of cognitive blind spots as an opportunity to update information, while the latter treats external challenge as an attack. Kahan (2012)'s ``identity-protective cognition'' theory explains the mechanism: when individuals feel their professional authority or intellectual image is threatened, they may unconsciously distort their reasoning to maintain their professional self-image.

\textbf{Positive mechanism.} When both B and C are oriented toward truth-seeking, logical challenges during dialogue (e.g., B pointing out a contradiction or reasoning gap in C's explanation) are treated as a means of increasing information rather than interpersonal offense. This psychological safety enables two key outcomes: first, B can ask deep, direct questions without excessive concern for hierarchical relationships; second, C can objectively display their hesitations, pauses, and revisions to initial judgments during complex decision-making. These preserved trial-and-error nodes are precisely the high-information-density material needed for post-training, because they reconstruct the complete arc of how the expert used experience to make decisions under uncertainty.

\textbf{Failure modes: two types of systemic data bias.} Without truth-seeking orientation, the dialogue produces data biases that are difficult to clean in post-processing. The two types share a similar underlying motivation (both are impression management) but manifest oppositely:

\begin{itemize}
\tightlist
\item
  \textbf{Failure Mode 1: Defensive bias.} When C is in a defensive mindset, facing B's deep questioning and encountering reasoning blocks may trigger a self-justification mechanism, producing so-called ``rationalization noise.'' Such arguments may be highly self-consistent on the surface but are in fact post-hoc defensive reasoning---not the actual causal chain C relied on when making the original judgment.
\item
  \textbf{Failure Mode 2: Performative bias.} This type is more insidious. When participants treat the dialogue as an occasion to display personal erudition, they may produce ``performative reasoning.'' Their output appears rich in information, but close examination reveals that the reasoning focus has shifted to peripheral topics that are convenient for demonstrating expertise, drifting away from an objective reconstruction of core decision logic.
\end{itemize}

\textbf{Cognitive compute appropriation: a unified explanatory framework for both biases.} Whether defensive or performative, the underlying mechanism of both biases can be unified as a cognitive resource allocation problem: participants divert limited cognitive bandwidth from ``reconstructing objective reasoning'' to ``maintaining social image.'' Cognitive load theory (Sweller, 1988---echoed in §3.4) classifies social image management as extraneous cognitive load, which directly crowds out the processing resources that should be directed to germane cognitive load (i.e., reasoning externalization). Self-regulatory resource theory (Baumeister et al., 1998) further notes that the executive function resources consumed by self-image management draw from the same finite pool as the resources required for deep reasoning. Experimental evidence from Schmader and Johns (2003) shows that when individuals feel their self-image is threatened, working memory capacity is significantly occupied, leading to degraded reasoning performance. This means that excluding defensive and arrogant personalities is not a moral judgment---it is an engineering necessity. It ensures that participants' full cognitive compute is invested in the core task of reasoning externalization, not consumed by the extraneous load of social image management.

\textbf{Chain reactions from adversarial interaction.} In dialogue, if one party deviates from truth-seeking, a chain reaction easily follows: a defensively minded expert inhibits the questioner's willingness to probe; an overly performative questioner may trigger defensiveness in the expert or cause them to withdraw from deep exchange. This non-constructive interaction pattern significantly degrades the signal-to-noise ratio of the data produced.

\subsubsection{Dimension 2: Epistemic Vigilance (Hard Constraint; Primarily Constraining B)}\label{dimension-2-epistemic-vigilance-hard-constraint-primarily-constraining-b}

Based on Sperber et al.~(2010)'s definition, ``epistemic vigilance'' is essentially the opposite of ``naïve gullibility.'' It requires that B must absolutely not be a passive receiver who ``accepts everything the expert says without question.'' B must maintain an instinctive professional skepticism. For readers from a computer science background without a cognitive science background, this concept can be concretely translated as: B's ability to sharply identify claims in C's output that ``appear logically consistent on the surface but are in fact reasoning-deficient.''

\textbf{Evolutionary foundations.} This ability has deeper theoretical roots. Mercier and Sperber (2011)`s ``Argumentative Theory of Reasoning'' proposes that human reasoning ability evolved not for solving problems alone but for evaluating and refuting weaknesses in others' arguments in social debate. In other words, identifying weak links in others' reasoning chains is a core capability optimized by natural selection in human cognitive architecture. B's role in the BC Protocol is precisely to activate and deploy this evolutionarily endowed evaluation mechanism---not to oppose the expert, but to use this natural ``argument-defect detector'' to locate the implicit reasoning nodes that have been omitted from C's output.

This is B's core required skill and the foundational mechanism for the effective operation of the BC Protocol. When expert C delivers a judgment, years of professional accumulation have internalized many intermediate reasoning steps as automated pattern recognition (the ``expert blind spot,'' Nathan \& Petrosino, 2003). To C, this direct inference---which omits the intermediate process---is self-evident. At this point, B must exercise epistemic vigilance: identify the implicit nodes that have been omitted from the logical chain and, through targeted follow-up questions, guide C to reconstruct the intuitive judgment as explicit logical steps.

Professionals trained in deep interviewing typically possess such capabilities. They can identify unstated premises and logical leaps from a respondent's seemingly coherent account.

This dimension is classified as a ``hard constraint'' because, without it, the BC Protocol's core elicitation mechanism cannot function. If B merely accepts information passively without identifying reasoning gaps, the follow-up questioning designed to fill knowledge gaps cannot be triggered. In this case, the dialogue may proceed smoothly on the surface but will mostly produce general conclusions lacking high-information-density reasoning processes (CoT). The dialogue effectively degrades into ordinary opinion collection.

In the cross-paper context of the 3+1 paper series, this division of labor has a precise information-theoretic position: B's epistemic vigilance primarily safeguards the three-party convergence legitimacy of the creative constraint \(Y\) (preventing \(Y\) from collapsing to only the ``creator's intent'' end), while C's expert intuition makes accurate (\(H(X|Y) \to 0\)) and surprising (\(H(X)\) elevated) dual-track judgments on creative choices \(X\) under legitimate \(Y\) (see Zou \& Xu, 2026, §1.2).

For convenience in subsequent references, we use the term \textbf{SNAKE mechanism} (Structured Narrative Analysis and Knowledge Elicitation) to refer to ``the targeted follow-up questioning and implicit knowledge externalization operations driven by B's epistemic vigilance within the dialogue.'' SNAKE is an internal term characterizing this micro-elicitation action. The highest-intensity form of SNAKE's questioning is \textbf{counterfactual probing}: rather than elaborating or confirming C's existing reasoning chain, it actively perturbs a premise variable that C has not explicitly stated but has implicitly assumed, forcing C to re-run the reasoning under a hypothetical premise contrary to reality. This mechanism and its implementation guidelines---designed not to disrupt the natural rhythm of dialogue---are detailed in §3.3.

\subsubsection{Dimension 3: Complementary Intelligence Profiles (Soft Constraint)}\label{dimension-3-complementary-intelligence-profiles-soft-constraint}

The BC Protocol is specifically designed for two-person pairings with different cognitive profiles. Drawing on Cattell (1963)'s distinction between ``fluid intelligence'' and ``crystallized intelligence,'' the ideal pairing has the elicitor (B) positioned at the high end of fluid intelligence (skilled at abstract reasoning, cross-domain analogy, and rapid pattern detection in unfamiliar contexts) while the domain expert (C) is positioned at the high end of crystallized intelligence (possessing extremely deep domain expertise accumulated through years of immersive experience).

This complementarity is mechanistically necessary. When two cognitively homogeneous peers engage in dialogue, as Wegner (1987) notes in his theory of Transactive Memory Systems, they form a highly interdependent cognitive division of labor and shared intuition. They tend to reach ``consensus'' quickly rather than ``externalizing.'' Because they share similar reasoning habits, both parties tend to take for granted the reasoning steps omitted from each other's arguments, making it difficult to trigger effective follow-up questions. Such dialogues proceed smoothly but produce low externalized information density, because the implicit premises both parties assume by default are precisely the critical reasoning chains most needed by post-training.

\textbf{Theoretical support for cognitive diversity.} Page (2007) in \emph{The Difference} argues that in problem-solving tasks, cognitively diverse teams systematically outperform homogeneous high-ability teams---diversity of perspectives and heuristics drives information externalization more than raw individual capability.

\textbf{The level at which questioning occurs: instinctive questioning vs.~within-framework questioning.} Page's and Wegner's theories can be further clarified through a micro-mechanism distinction in the BC Protocol: distinguishing the \textbf{level} at which questioning occurs rather than its frequency. In dialogue between cognitively homogeneous peers, even with dense questioning, questions almost entirely fall \textbf{within} the conceptual framework the expert has established---``What else within this framework has not been explained?''---this is a dutiful, compliant form of within-framework questioning. It can fill in the blanks of the existing framework but cannot touch the framework's own boundaries. B at the high end of fluid intelligence tends toward a different form of questioning: doubting the conceptual framework the expert has established in the first place---``Why this framework, and not another?''---this is instinctive framework-level questioning. Each such question has some probability of pulling the expert out of automated pattern recognition, forcing C to re-examine framework assumptions that have long since sunk below the threshold of consciousness.

It is worth noting that this instinctive questioning does not come from B trying harder or undergoing more conscious training. It is a byproduct of B's cognitive structure: precisely because B has not been long immersed in the domain's conceptual framework the way C has, B can stand outside the framework and ask. In other words, for the specific elicitation task required by the BC Protocol, the fluid intelligence practitioner's knowledge disadvantage relative to domain peers is itself the source of their questioning-level advantage. This is the micro-causal mechanism underlying the cognitive complementarity this section advocates---B and C are not interchangeable, not only because they hold different stocks of knowledge, but because their questioning instincts fall naturally at different levels.

By contrast, the complementarity between fluid and crystallized intelligence creates effective cognitive difference. B's cross-domain questioning perspective touches the logical blind spots that C is less consciously inclined to articulate within their domain habits; C's domain experience can answer the professional judgments that B cannot derive by logic alone. The cognitive gap between the two provides the necessary exploratory space and questioning motivation for eliciting implicit knowledge.

\subsubsection{Dimension 4: Calibrated Cognitive Gap (Soft Constraint)}\label{dimension-4-calibrated-cognitive-gap-soft-constraint}

The knowledge gap between B and C must be precisely calibrated---neither too large nor too small. Drawing on Vygotsky (1978)'s concept of the Zone of Proximal Development (ZPD), B should be positioned just at the ``instructional edge'' of C's ZPD: close enough to understand C's domain jargon in real time and keep pace with C's thinking; but far enough that, at truly critical judgment nodes, B experiences a genuine cognitive gap.

\textbf{Theoretical support for ``appropriate difficulty.''} Bjork (1994)'s concept of ``desirable difficulties'' shows that moderate cognitive resistance actually promotes deeper information externalization. In the context of the BC Protocol, B's appropriately calibrated ``lack of understanding'' of C constitutes a desirable difficulty---forcing C to ``reverse-decompress'' automated pattern recognition into explicit reasoning steps.

If B's knowledge base is insufficient (a complete layperson), their questions tend to be too broad (e.g., ``Can you explain why this passage is good?''), easily guiding C into a basic ``popular science teaching mode'' rather than the ``reasoning externalization mode'' research requires. Conversely, if B's knowledge base is nearly equivalent to C's (a peer), the high volume of implicit premises both parties share significantly reduces questioning frequency. The ideal state is: B can fully understand C's content, but at critical decision nodes maintains sharp questioning awareness of ``how C arrived at this judgment.''

\subsubsection{Dimension 5: Calibrated Ignorance (Soft Constraint)}\label{dimension-5-calibrated-ignorance-soft-constraint}

We define ``calibrated ignorance'' as a metacognitive state of ``knowing precisely what one does not know.'' It follows from Tetlock (2005)'s emphasis on ``calibration'' in expert judgment research, while differing from the philosophical tradition of ``docta ignorantia'' (learned ignorance): the latter is a philosophical stance, while the former is a practically operational cognitive state.

\textbf{Theoretical support from curiosity and metacognition.} Loewenstein (1994)'s Information Gap Theory holds that high-quality questioning arises when individuals precisely perceive the boundary between what they know and what they do not---when the boundary is fuzzy, curiosity cannot be activated; when the boundary is clear, questions naturally point toward the direction of maximum information gain.

In BC Protocol dialogue, calibrated ignorance mainly manifests as: B can direct questions precisely toward the information gap between ``what C knows but has not yet stated''---i.e., the interval between C's cognition and their verbal output. Calibrated questioning maximizes information gain by pointing directly at the implicit premises C habitually skips. If B's ignorance is unverified and uncalibrated, their questions may be redundant (repeating already-given information), or may point toward blind spots beyond C's expertise (leading to pointless speculation or invalid answers).

\textbf{Domain-dependent threshold model}

The knowledge baseline required for ``calibrated ignorance'' is not a fixed constant---it rises and falls dynamically with the knowledge architecture of the target domain. The yardstick for distinguishing different domain types is the \textbf{applicability of first principles}: if a layperson can infer and judge logical correctness and quality using only first principles, this is a high-explicitness domain. Conversely, if knowledge and high-order judgment fundamentally cannot be derived through first principles, this is a high-tacit-knowledge domain.

\begin{itemize}
\tightlist
\item
  \textbf{High-explicitness domains (e.g., physics, engineering, computer science):} B needs a \textbf{lower} background threshold. Knowledge in these domains has a structured hierarchical distribution with strong inferability. Even with no familiarity with specific physical parameters, basic scientific literacy allows B to easily detect whether ``because X, therefore Y'' shows a break in the logical chain.
\item
  \textbf{High-tacit-knowledge domains (e.g., narrative fiction, film aesthetics, design):} B needs a \textbf{higher} background threshold. Expert judgment in such domains is heavily dependent on pattern matching and intuition formed through long immersion. If B does not have a rich aesthetic intuition as a foundation, when C directly delivers a high-order conclusion while skipping ten thousand micro-perceptual steps, B will not even know what happened---because B has no perceptual ``vocabulary'' to even recognize that ``a link was skipped here.''
\item
  \textbf{Institutionally mixed domains (e.g., law, finance, medicine):} B needs a \textbf{moderate} threshold. These domains are half explicit strong rules and half implicit flexible judgment. B needs to master basic jargon and rule logic in order to distinguish which part of C's output is ``rote recitation'' with no information value, and which part is the truly high-order discretionary judgment that relies on implicit experience.
\end{itemize}

This can be condensed into a core formula: \textbf{the optimal knowledge entry threshold B requires is a function of the degree of tacit knowledge in the domain.} The more domain knowledge lies in the dark and is difficult to articulate, the higher B's initial position must be. The more explicit and structured the domain knowledge, the lower the starting point from which B can launch precise questioning.

\subsubsection{Dimension 6: Complementary Information Processing Preferences (Soft Constraint)}\label{dimension-6-complementary-information-processing-preferences-soft-constraint}

Effective knowledge elicitation benefits from a complementary tension between B and C in their information processing styles. In the ideal configuration, B typically has a stronger \textbf{framing bias} (trust-and-frame bias): a tendency to quickly build a preliminary conceptual framework based on C's output. C, by contrast, has a stronger \textbf{deepening bias} (challenge-and-deepen bias): a tendency to examine and correct the simplified framework B has built, adding more complex micro-details and professional constraints.

\textbf{Theoretical support from constructive controversy.} Johnson and Johnson (2009)'s Constructive Controversy theory provides direct theoretical support: when both parties hold different positions but share a common goal (joint truth-seeking), the ``frame---challenge---revise'' cycle systematically produces higher-quality cognitive output than either party thinking alone. B's framework-building and C's professional correction in the BC Protocol are precisely micro-instances of this constructive controversy.

Without B's framework summarization, dialogue easily loses focus. Without C's detail correction, conclusions may remain shallow. It is precisely in the interplay of ``framing and correcting,'' ``breadth and depth,'' that high-quality CoT samples with complete logical structure and rich detail are produced.

\begin{figure}[t]
  \centering
  \includegraphics[width=0.96\linewidth]{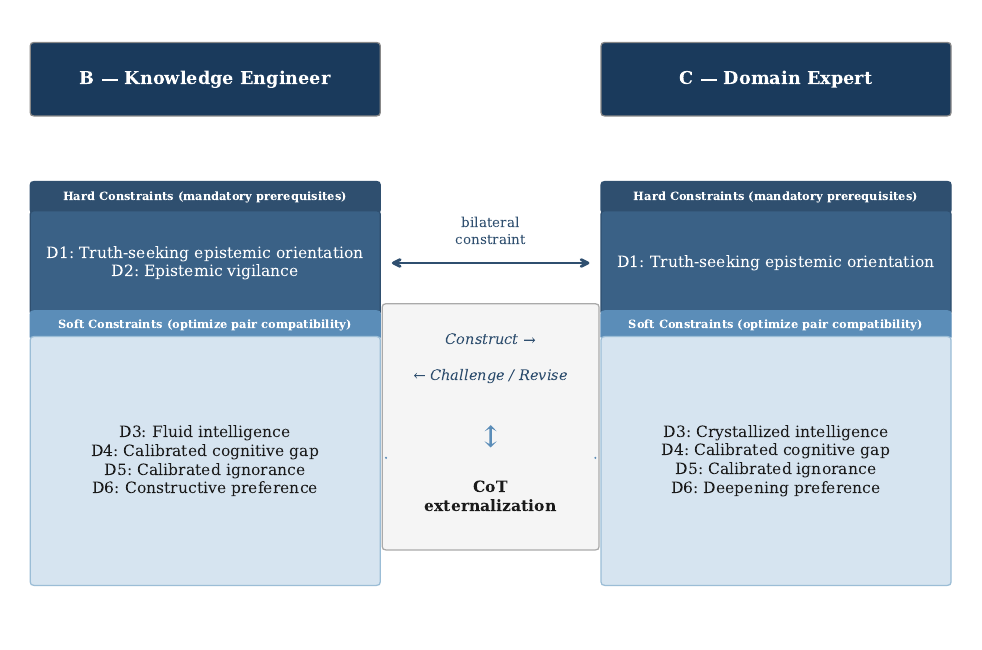}
  \caption{\textbf{The Participant Aptitude Model.} Left: knowledge engineer (B); right: domain expert (C). Hard constraints (dark shading) are non-negotiable prerequisites for both roles. Soft constraints optimize pair compatibility. The central loop represents the construct\textendash challenge\textendash revise cycle that drives CoT externalization.}
  \label{fig:aptitude}
\end{figure}

\subsection{Dialogue Facilitation Guide: No Mechanical Templates}\label{dialogue-facilitation-guide-no-mechanical-templates}

\textbf{Core design principle: Selection-over-Prescription.} The BC Protocol deliberately specifies no rigid questioning routines or standardized script templates. This is not methodological laziness---it is a core design decision grounded in theory:

First, \textbf{internal theoretical consistency.} The paper's own core argument is that tacit knowledge cannot be fully externalized through procedural means. If we then try to proceduralize the elicitation process itself into a rigid script, we contradict ourselves at the methodological level. High-quality knowledge elicitation is itself a high-order skill that depends on B's implicit professional capabilities and equally cannot be reduced to a standardized operating manual.

Second, \textbf{operational logic.} As long as the right people have been rigorously selected at the front end (§3.2 Participant Aptitude Model), those with epistemic vigilance and truth-seeking orientation already know intuitively how to follow up, decompose, and probe during dialogue. Forcing a questioning script would suppress their professional intuition. Experienced knowledge elicitors typically have personalized methods and interviewing techniques. In the face of dynamic dialogue, imposing a standardized script may actually limit elicitation effectiveness.

Third, \textbf{shifting the locus of quality assurance upstream.} The BC Protocol moves the center of quality control from process regulation to personnel selection. The Participant Aptitude Model is the quality assurance mechanism for the entire protocol. Traditional knowledge engineering methodology tries to compensate for mediocre practitioners through exhaustive process design. The BC Protocol trades rigorous practitioner selection for maximum procedural freedom.

Therefore, B's core guiding principle in practice is to remain oriented toward truth-seeking at all times: relying on professional intuition, flexibly applying personal experience for deep follow-up questioning. Reducing formal constraints helps more completely and systematically draw out the reasoning processes the expert has implicitly internalized.

\textbf{Dialogue termination: B controls the exit signal.} Following the same logical chain as the ``no mechanical templates'' principle above is the question of when to terminate dialogue on a single creative decision. We explicitly assign this authority to B, not to C. The reason was already laid out in Dimension 2 (§3.2.4): as an expert deeply immersed in their domain, C is naturally subject to the ``expert blind spot'' regarding the implicit nodes in their own judgments that have been absorbed into automated pattern recognition (Nathan \& Petrosino, 2003). If C were to self-judge ``this one is done,'' their exit threshold would be tied to how self-evident the judgment feels to them---the more obvious the judgment is to C, the earlier they would stop. This degradation path would cause BC dialogue to quietly revert, in the tail of each sample, to the ``expert solo writing'' mode diagnosed in §2: surface interaction continues, but the deeper implicit nodes remain unexposed due to C's blind spot.

B's exit signal, by contrast, is unrelated to the ``obviousness'' of the judgment. It is tied to ``whether uncertainty has been exhausted'': B enters a terminal state only when epistemic vigilance scanning across the entire dialogue no longer triggers new follow-up questions, and further questioning would only prompt C to produce rationalization noise (§3.2.3 Failure Mode 1) rather than adding new reasoning nodes. From this it follows that the actual length of a single CoT is determined by the genuine complexity of that judgment---not by any ``upper limit on dialogue turns'' or ``full coverage of dimensions'' or other explicit formal conditions. CoT samples may therefore be significantly longer than conventional alignment data samples. Notably, this is an expected methodological outcome. It is handled at the post-training end by engineering capacity (multi-GPU parallelism, long-context support), and should not be rectified by any compression operation at the dialogue layer that would constrain the natural output form.

\textbf{B's exit judgment as an implicit bottleneck.} The above design reveals an inherent cost of the Selection-over-Prescription principle: when the locus of quality assurance is moved upstream from process regulation to personnel selection (§3.3 third point), B's exit judgment itself becomes an implicit bottleneck for data quality---one that does not reduce to any explicit protocol. The precision of B's exit timing on each sample directly determines that sample's effective information density: exit one step too early and the implicit node remains unexposed within C's blind spot; exit one step too late and C shifts into rationalization, injecting false reasoning. This means that, given a batch of CoT samples produced from BC dialogues, their methodological quality cannot be estimated solely by dialogue count or sample character length---it requires considering the distributional quality of B's epistemic vigilance at each sample's exit state. This is the methodological cost that the BC Protocol necessarily incurs by choosing not to compensate for practitioner quality through explicit processes (§3.3 third point). It is not a remediable engineering defect.

\subsubsection{Counterfactual Probing: The Highest-Intensity Form of Epistemic Vigilance}\label{counterfactual-probing-the-highest-intensity-form-of-epistemic-vigilance}

§3.3 consistently argues for ``no mechanical templates.'' But this does not preclude an epistemological characterization of B's questioning forms by level---some questioning forms are structurally better at drawing out C's implicit causal model. We explicitly mark the highest-intensity category here so that the post-processing stage (§3.5) can grade CoT samples by quality.

\textbf{Concept.} Borrowing the ``counterfactual'' concept from causal inference (Lewis, 1973; Pearl, 2009), \textbf{counterfactual probing} refers to a B-initiated questioning action that does not elaborate on or confirm C's existing reasoning chain. Instead, it \textbf{actively perturbs a premise variable that C has not explicitly stated but has implicitly assumed in the reasoning}, forcing C to re-run the reasoning under a hypothetical premise contrary to reality. Its form is: ``If it had not been X but X' instead, what would Y become?''---the key is that the perturbed X must be an implicit premise C has not explicitly stated but has silently imported into the reasoning, not a restatement of facts C has already discussed.

\textbf{Why this does not violate ``no mechanical templates.''} Counterfactual probing is not a proceduralized action that ``must be mechanically triggered after every reasoning segment.'' It is a perturbation that B selectively inserts---during a semantic pause or breath between segments---after sensing that C has completed a natural reasoning segment containing a load-bearing claim. Whether to insert, where to insert, and which variable to perturb are all judged in real time by B's epistemic vigilance; no explicit rule governs them. This form is therefore the highest-intensity concrete realization of the SNAKE mechanism (§3.2.4), not a separately mandated script.

\textbf{Why it contributes most to data quality.} In an information-theoretic framework (see the companion paper \emph{Calibrated Surprise}), a statement's causal load is positively correlated with the uncertainty it converges in the conditional entropy \(H(X|Y)\). A reasoning segment that maintains its conclusion even after counterfactual perturbation means \(H(X|Y)\) has genuinely been narrowed in that direction. A reasoning segment whose conclusion changes after perturbation makes visible one branch of the ``tree of options that was cut''---a branch inside C's mind---as a causal signal that LLMs can learn from. Both types of output are scarce in public corpora: the former because humans rarely add counterfactual validity proofs to already-established causal chains, the latter because rejected options are almost never actively externalized by their authors. Counterfactual probing is the lowest-cost entry point for simultaneously eliciting both types of signal.

\textbf{Implementation guidelines (guidance, not a script):}

\begin{itemize}
\tightlist
\item
  \textbf{Perturb implicit premises; do not restate stated premises.} If C has already explained ``because X therefore Y,'' then asking ``if not X then not Y, right?'' is merely a confirmatory restatement with limited information gain. Instead, perturb an adjacent condition that C has not stated but that their reasoning implicitly requires.
\item
  \textbf{Insert at natural pauses; do not interrupt mid-reasoning.} Counterfactual probing is released only after C has completed a full logical closure, not inserted mid-narrative flow, to avoid damaging the advantage of the voice medium in preserving trial-and-error nodes (§3.4).
\item
  \textbf{At most once per reasoning segment, constrained by load-bearing claim density.} If a segment contains only one load-bearing claim, insert only one counterfactual probe. If the reasoning is purely descriptive, aesthetic, or a standalone setting, do not insert. Detection method: apply a reversal test to the causal connectives in the segment (``because/therefore/must/otherwise''). If the segment still holds after reversal, the connective is not load-bearing; if the segment collapses, it is load-bearing.
\item
  \textbf{Both decoy and open forms can be used.} The decoy form offers a deliberately incomplete or misdirected counterfactual inference, forcing C to actively correct it. The open form raises the perturbation without offering an inference, letting C freely develop a new branch. The two forms correspond to different types of load-bearing claims and are switched in real time by B according to dialogue rhythm.
\end{itemize}

\textbf{Fundamental difference from ``post-hoc counterfactual data augmentation.''} The BC Protocol moves the counterfactual elicitation action upstream and embeds it within the dialogue itself, so each dialogue segment is collected with its counterfactual branches already present. This is structurally different from the offline counterfactual rewriting path currently used in industry (Kaushik et al., 2020). The cost analysis of this difference and the formal definition of the counterfactual density metric are detailed in §6.4.

\subsection{Voice vs.~Text as Elicitation Medium: Theoretical Rationale}\label{voice-vs.-text-as-elicitation-medium-theoretical-rationale}

The BC Protocol mandates \textbf{dual-person voice dialogue} (rather than written text) as the medium for implicit knowledge elicitation. This is not a pragmatic compromise---it is a principled rule established to prevent \textbf{lossy self-editing}.

Human beings have long written professional text with the goal of making it ``easy for others to read'' or ``conveying a clear conclusion.'' The more senior an expert, the more mature their ``muscle memory'' for editing, formatting, and polishing---all driven by reader experience, logical coherence, and professional image maintenance. When they try to output judgments in writing, their goal instinctively orients toward producing a clean, flawless conclusive statement. This implicit pre-publication editing process is directly at odds with our ultimate purpose of extracting native reasoning chains (CoT) for machine post-training: in order to output clean conclusions, experts consciously delete as waste the very intermediate steps that involve leaps, iterations, self-reversals, and trial and error. Text must therefore never be relied upon to elicit deep implicit reasoning.

The voice medium effectively bypasses this written-polish filter, because the sequence of oral expression is closer to the original state of cognition as it unfolds. Even single-person self-narration recorded on audio preserves significantly more trial-and-error nodes than written text. The ``two-person real-time voice dialogue'' specified by the BC Protocol works even better: the immediacy of real-time interaction prevents the expert from constructing a coherent, self-consistent monologue in advance, thereby prompting the expert to naturally surface their initial reasoning path when responding to follow-up questions. Verbal self-corrections and semantic pauses---such as ``wait, that characterization was imprecise'' or ``actually, I need to consider another scenario''---are often treated as redundant information in ordinary communication. In this protocol, however, they are extremely scarce yet highly information-dense data nodes for distilling high-fidelity reasoning processes.

This design also aligns with \textbf{cognitive load theory} (Sweller, 1988). Attending to written vocabulary and grammatical structure consumes large amounts of expert cognitive bandwidth. Pure voice exchange requires no attention to such formatting concerns, freeing that computational capacity for reasoning through the core problem. Finally, voice preserves rich \textbf{prosodic cues}. In which sentence is the expert highly confident? In which supplementary remark does a slight waver appear at the trailing syllable? These features signal to B exactly where to probe next for deeper follow-up questioning. This multi-dimensional interactive feedback simply does not exist in plain text exchange.

\subsection{Post-Processing}\label{post-processing}

The core task of the post-processing stage is to transform raw dialogue audio into natural language chain-of-thought (CoT) samples meeting LLM post-training standards. This workflow typically includes: high-fidelity transcription of voice audio that preserves hesitations and verbal slips; segmentation of lengthy dialogue into individual judgment units that are each focused and self-contained; filtering out fragments with incomplete reasoning chains or logical gaps; and finally removing irrelevant small talk and auxiliary follow-up questions, formatting the remainder into coherent reasoning paragraphs.

The guiding principle throughout this stage is ``information preservation.'' Processing work should remove irrelevant noise while completely preserving the expert's characteristic features during reasoning---trial and error, self-correction, and hesitation. Compared to the single correct conclusion ultimately reached, these intermediate self-corrections carry higher information gain for LLM training and more faithfully reflect the genuine reasoning process underlying expert judgment.

\begin{figure}[t]
  \centering
  \includegraphics[width=0.96\linewidth]{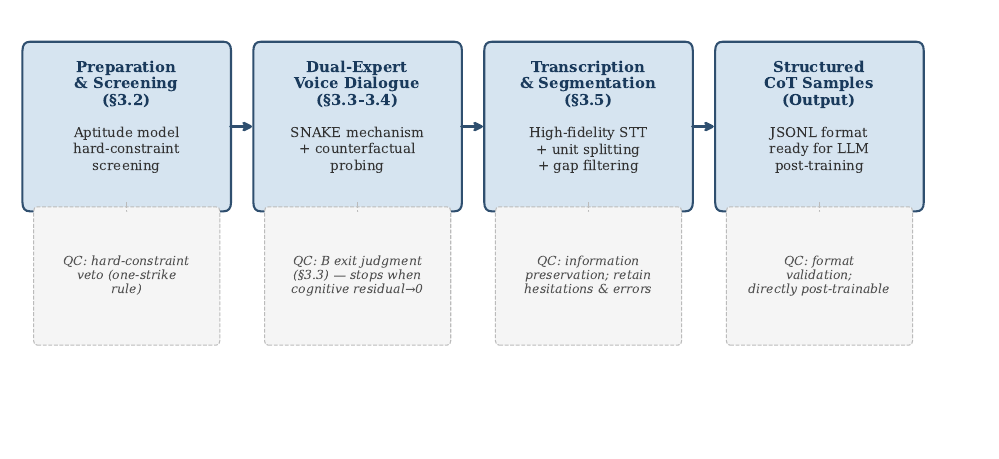}
  \caption{\textbf{End-to-end BC Protocol pipeline.} Each stage has explicit quality-control checkpoints (dashed boxes). The output is structured CoT samples in JSONL format, directly usable for LLM post-training.}
  \label{fig:pipeline}
\end{figure}

\section{Empirical Validation}\label{empirical-validation}

\begin{quote}
This section focuses on \textbf{comparative validation of intrinsic data quality}. Validation of downstream fine-tuning effects is left to the CQA companion paper.
\end{quote}

\subsection{Experimental Domain and Materials}\label{experimental-domain-and-materials}

This experiment uses \textbf{narrative fiction}---specifically, creative judgments in the genres of social detective fiction and Japanese youth mystery (e.g., the \emph{Hyouka} series, Japanese suspense, Jin Yong-style character arrangements)---as evaluation material. This domain was chosen for three reasons: (1) Narrative judgment falls within the \textbf{high-tacit-knowledge domain} as defined in §3.2.7. Expert inference in this domain depends heavily on pattern matching formed through long immersion and cannot be derived through first principles---making it the most stringent test of the ``fluid--crystallized intelligence pairing'' claimed by the BC Protocol. (2) Creative judgment has no unique correct answer, forcing evaluation to focus on \textbf{the quality of the reasoning process itself} rather than conclusion correctness---precisely what CoT data quality assessment requires. (3) The team (B \& C) has practical expertise in this domain, allowing the Expert Solo control group (Group B) to be independently completed by the same C who genuinely has expert-level judgment, ruling out external criticism that ``the baseline was deliberately weakened.''

\subsection{Control Group Design}\label{control-group-design}

This experiment includes two direct comparison groups. \textbf{Compared to the 4-group design listed in an earlier draft of the paper (which included a commercial SOTA reference baseline group and a crowdsourcing group), we narrowed the design to 2 groups in the 0512 experimental plan, removing the commercial API baseline group and the crowdsourcing group.} The former cannot constitute a credible baseline in high-tacit-knowledge domains like narrative fiction (even top commercial models exhibit severe ``post-hoc rationalization'' tendencies in this domain). The latter's elicitation ceiling has already been structurally refuted by the failure mode analysis in §2.1.1---including it as a comparison would consume experimental budget without contributing interpretable variance.

{\def\LTcaptype{none} % do not increment counter
\begin{longtable}[]{@{}
  >{\raggedright\arraybackslash}p{(\linewidth - 6\tabcolsep) * \real{0.2500}}
  >{\raggedright\arraybackslash}p{(\linewidth - 6\tabcolsep) * \real{0.2500}}
  >{\raggedright\arraybackslash}p{(\linewidth - 6\tabcolsep) * \real{0.2500}}
  >{\raggedright\arraybackslash}p{(\linewidth - 6\tabcolsep) * \real{0.2500}}@{}}
\toprule\noalign{}
\begin{minipage}[b]{\linewidth}\raggedright
Group
\end{minipage} & \begin{minipage}[b]{\linewidth}\raggedright
Data Production Method
\end{minipage} & \begin{minipage}[b]{\linewidth}\raggedright
Sample Size
\end{minipage} & \begin{minipage}[b]{\linewidth}\raggedright
Role
\end{minipage} \\
\midrule\noalign{}
\endhead
\bottomrule\noalign{}
\endlastfoot
\textbf{Group A (BC Protocol experimental group)} & B and C structured dual-person voice dialogue → high-fidelity transcription → §3.5 post-processing & 20 & Core experimental method \\
\textbf{Group B (Expert Solo control group)} & Same C, without any contact with Group A's dialogue, independently writes CoT on \textbf{the same set of creative topics} & 20 & Direct baseline \\
\end{longtable}
}

\begin{quote}
\textbf{Sequential contamination prevention.} Group B's solo writing must be completed before Group A's dialogue is compiled, to prevent C's solo version from being contaminated by B's questioning paths in the dialogue. Specific implementation details are in the supplementary topic list.
\end{quote}

Both groups use a unified JSONL format with fields \texttt{\{id,\ group,\ preamble,\ cot\_body,\ summary,\ metadata\}}, archived in the supplementary data files (\texttt{group\_A.jsonl} and \texttt{group\_B.jsonl}). The \texttt{preamble} (background context) is provided solely as background for the review LLM to understand the CoT context and \textbf{is not scored}---this rule is explicitly declared in all review prompts.

\subsection{Evaluation Method: Metric Dimensions and Blind Assessment Design}\label{evaluation-method-metric-dimensions-and-blind-assessment-design}

\subsubsection{Five-Dimension Quality Metric System for CoT}\label{five-dimension-quality-metric-system-for-cot}

This paper uses \textbf{multi-LLM consensus review} rather than human blind assessment, for three reasons: (1) The cognitive load required to score each CoT across five dimensions in detail is enormous; human reviewer fatigue and drift would introduce variance that is difficult to control. (2) LLM review consistency and reproducibility are significantly higher than human review, and the review prompt itself can be published as an appendix, giving the entire evaluation \textbf{protocol-level reproducibility}---something no human review can provide. (3) By simultaneously using flagship models from three vendors (avoiding single-model bias) and reporting cross-model agreement (Krippendorff's \(\alpha\)), the design constitutes a multi-reviewer setup with built-in cross-validation.

{\def\LTcaptype{none} % do not increment counter
\begin{longtable}[]{@{}
  >{\raggedright\arraybackslash}p{(\linewidth - 4\tabcolsep) * \real{0.3333}}
  >{\raggedright\arraybackslash}p{(\linewidth - 4\tabcolsep) * \real{0.3333}}
  >{\raggedright\arraybackslash}p{(\linewidth - 4\tabcolsep) * \real{0.3333}}@{}}
\toprule\noalign{}
\begin{minipage}[b]{\linewidth}\raggedright
Dimension
\end{minipage} & \begin{minipage}[b]{\linewidth}\raggedright
Operational Definition
\end{minipage} & \begin{minipage}[b]{\linewidth}\raggedright
Scoring Anchors
\end{minipage} \\
\midrule\noalign{}
\endhead
\bottomrule\noalign{}
\endlastfoot
\textbf{D1: Reasoning Chain Completeness} & Are there missing links in the derivation from premises to conclusion? & 1=many logical leaps; 3=backbone complete but some details missing; 5=every reasoning step is externalized, seamlessly connected \\
\textbf{D2: Implicit Premise Externalization Rate} & Are reasoning premises ``typically taken as self-evident by experts'' made explicit? & 1=almost all implicit premises are assumed; 3=core implicit premises are externalized; 5=even micro-premises underlying deep aesthetic intuition are verbalized \\
\textbf{D3: Naturalness of Reasoning Process} & Does the output preserve characteristic process features such as trial and error, hesitation, and self-correction from the expert's original reasoning? & 1=flat, polished conclusion stacking with no trace of trial and error; 3=occasional self-correction; 5=a live, breathing reasoning flow \\
\textbf{D4: Information Density} & The amount of non-deletable independent reasoning information per unit length & 1=much filler/restatement; 3=informative but with redundancy; 5=every sentence is non-deletable \\
\textbf{D5: Counterfactual Density} (§3.3.1 / §6.4 core contribution) & The ratio of load-bearing claims that are explicitly probed by counterfactual perturbation & 1=zero counterfactual branches; 3=moderate coverage of load-bearing claims; 5=every load-bearing claim is perturbed and a new branch is traced \\
\end{longtable}
}

\subsubsection{Review Pipeline and Blinding Design}\label{review-pipeline-and-blinding-design}

\textbf{Review models.} We use flagship models from three vendors:

{\def\LTcaptype{none} % do not increment counter
\begin{longtable}[]{@{}lll@{}}
\toprule\noalign{}
Label & Model ID & Vendor \\
\midrule\noalign{}
\endhead
\bottomrule\noalign{}
\endlastfoot
\texttt{gpt\_4o} & \texttt{gpt-4o} & OpenAI \\
\texttt{claude\_opus} & \texttt{claude-opus-4-5} & Anthropic \\
\texttt{gemini\_pro} & \texttt{gemini-2.5-pro} & Google \\
\end{longtable}
}

All API calls are routed through a proxy endpoint (\texttt{api.apiyi.com}), which remained stable throughout. Aside from Claude Opus requiring up to 3 JSON parse retries on isolated samples due to unescaped quotes in its output, the remaining 597 calls succeeded on the first attempt.

\textbf{Blinding.} Each review prompt contains no group labels such as ``Group A,'' ``Group B,'' ``BC dialogue,'' or ``Expert Solo.'' Only \texttt{\{preamble,\ cot\_body\}} and the scoring rubric for that dimension are provided. The review model has no information about the sample's origin when scoring.

\textbf{Evaluation scale.} \(20 \times 2\) (samples) \(\times 5\) (dimensions) \(\times 3\) (models) \(= 600\) independent scoring calls, each producing a JSON object \texttt{\{"score":\ 1-5,\ "rationale":\ "..."\}}. Raw records are archived in the supplementary evaluation archive, with one jsonl file per \texttt{(model,\ dimension)} pair.

\subsubsection{Statistical Tests}\label{statistical-tests}

For each CoT × dimension pair, scores from the three review models are first averaged to yield a 40 × 5 matrix. Between-group comparison uses the \textbf{Mann-Whitney U test} (non-parametric, since 1--5 scores are ordinal) and reports \textbf{Cliff's \(\delta\)} as the effect size. Cross-model agreement is reported as \textbf{Krippendorff's \(\alpha\)}.

Significance threshold: \(\alpha = 0.05\).

\subsection{Statement on Connection to Fine-Tuning Experiments}\label{statement-on-connection-to-fine-tuning-experiments}

This paper does not include fine-tuning experiments. Validation of the data's downstream fine-tuning effects is handled by the companion CQA paper. This paper focuses on comparing the methodological merits of the data production approach itself---standard practice for data engineering methodology papers. Zhou et al.~(2023)'s LIMA likewise treats ``data quality argumentation'' and ``downstream fine-tuning validation'' as two separate layers of contribution across different publications.

\section{Results and Analysis}\label{results-and-analysis}

\subsection{Main Results Table}\label{main-results-table}

Table 1 reports, for each of the five dimensions, the mean, median, standard deviation, Mann-Whitney U test \(p\)-value, Cliff's \(\delta\) effect size, and Krippendorff's \(\alpha\) inter-model agreement for Group A (BC Protocol, \(n=20\)) vs.~Group B (Expert Solo, \(n=20\)).

\textbf{Table 1. BC Protocol (Group A) vs.~Expert Solo (Group B): Five-Dimension Blind Assessment Results}

{\def\LTcaptype{none} % do not increment counter
\begin{longtable}[]{@{}
  >{\raggedright\arraybackslash}p{(\linewidth - 14\tabcolsep) * \real{0.1250}}
  >{\centering\arraybackslash}p{(\linewidth - 14\tabcolsep) * \real{0.1250}}
  >{\centering\arraybackslash}p{(\linewidth - 14\tabcolsep) * \real{0.1250}}
  >{\centering\arraybackslash}p{(\linewidth - 14\tabcolsep) * \real{0.1250}}
  >{\centering\arraybackslash}p{(\linewidth - 14\tabcolsep) * \real{0.1250}}
  >{\centering\arraybackslash}p{(\linewidth - 14\tabcolsep) * \real{0.1250}}
  >{\centering\arraybackslash}p{(\linewidth - 14\tabcolsep) * \real{0.1250}}
  >{\centering\arraybackslash}p{(\linewidth - 14\tabcolsep) * \real{0.1250}}@{}}
\toprule\noalign{}
\begin{minipage}[b]{\linewidth}\raggedright
Dimension
\end{minipage} & \begin{minipage}[b]{\linewidth}\centering
A Mean
\end{minipage} & \begin{minipage}[b]{\linewidth}\centering
A Median
\end{minipage} & \begin{minipage}[b]{\linewidth}\centering
B Mean
\end{minipage} & \begin{minipage}[b]{\linewidth}\centering
B Median
\end{minipage} & \begin{minipage}[b]{\linewidth}\centering
\(p\)
\end{minipage} & \begin{minipage}[b]{\linewidth}\centering
Cliff's \(\delta\)
\end{minipage} & \begin{minipage}[b]{\linewidth}\centering
Krippendorff \(\alpha\)
\end{minipage} \\
\midrule\noalign{}
\endhead
\bottomrule\noalign{}
\endlastfoot
D1: Reasoning Chain Completeness & 2.91 & 3.00 & 2.63 & 2.50 & 0.256 & +0.21 & $-$0.01 \\
D2: Implicit Premise Externalization Rate & 2.85 & 2.67 & 2.88 & 3.00 & 0.859 & $-$0.04 & \textbf{+0.38} \\
\textbf{D3: Naturalness of Reasoning Process} & \textbf{4.80} & \textbf{5.00} & \textbf{1.30} & \textbf{1.33} & \textbf{\(\mathbf{2.4\times 10^{-8}}\)} & \textbf{+1.00} & $-$0.06 \\
D4: Information Density & 3.02 & 3.00 & \textbf{4.07} & \textbf{4.33} & \textbf{\(\mathbf{1.1\times 10^{-4}}\)} & \textbf{$-$0.73} & $-$0.14 \\
D5: Counterfactual Density & 1.50 & 1.33 & 1.20 & 1.00 & 0.101 & +0.27 & $-$0.03 \\
\end{longtable}
}

\begin{quote}
\textbf{Sample size note:} On D4 (Information Density), 2 Group A samples were excluded due to incomplete JSON output from the review model, so \(n_A=18\). All other dimensions have \(n_A=n_B=20\).
\textbf{Raw data:} See the supplementary aggregated results; box plots in the supplementary box plot figures.
\end{quote}

\begin{figure}[t]
  \centering
  \includegraphics[width=0.96\linewidth]{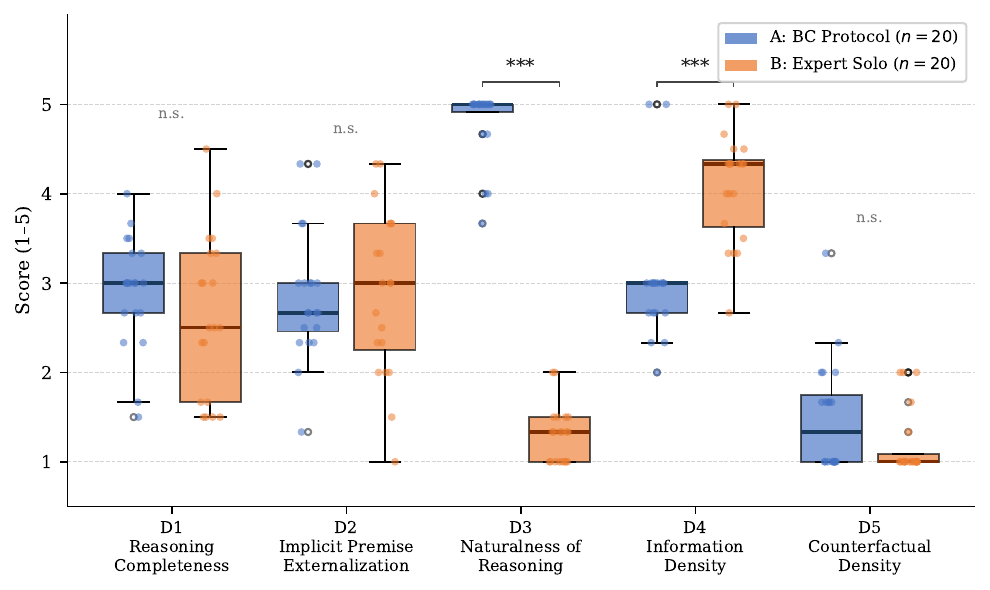}
  \caption{\textbf{Grouped box plots of blind evaluation scores across five dimensions.}
    A~=~BC Protocol ($n=20$, blue) vs.\ B~=~Expert Solo ($n=20$, orange).
    Each score is the mean of three judge models (GPT-4o, Claude Opus~4.5, Gemini~2.5~Pro).
    Jittered strip plots show individual sample distributions.
    Significance brackets: $***$~$p < 0.001$ (Mann-Whitney $U$ test, two-sided).
    D4 note: $n_A=18$ due to two incomplete judge outputs.}
  \label{fig:boxplot}
\end{figure}

\subsection{Three Layered Conclusions}\label{three-layered-conclusions}

The main results have a highly bifurcated structure. We organize them into three categories.

\subsubsection{Overwhelming Advantage: D3 Naturalness of Reasoning Process}\label{overwhelming-advantage-d3-naturalness-of-reasoning-process}

This is the most definitive finding of the experiment. Across 600 blind ratings, the three review models produced a split on this dimension that is \textbf{almost impossible to attribute to chance}: all 20 Group A CoT samples concentrated in the 4.67--5.00 range, and all 20 Group B samples in the 1.00--2.00 range. Cliff's \(\delta = 1.00\) (the theoretical maximum) means that every single Group A sample strictly outscores every single Group B sample on this dimension. \(p = 2.4 \times 10^{-8}\) far exceeds any common significance threshold.

This result directly validates the core theoretical claim in §3.4: \textbf{when writing, experts instinctively delete trial-and-error nodes, hesitation markers, and directional switches in pursuit of clean, conclusive output}. The BC dual dialogue, through the cognitive rhythm of real-time interaction, structurally preserves these ``live reasoning features.'' All three review models consistently identified: Group A reasoning chains have a sense of breath---self-correction, iterative convergence and divergence; Group B reasoning chains are flat, polished stacking of conclusions with almost no detectable trace of the original reasoning as it unfolded.

\begin{quote}
\textbf{Methodological note on D3's Krippendorff \(\alpha = -0.06\).} This negative value does not indicate model disagreement. On the contrary---it is a statistical artifact of \textbf{near-zero within-group variance} (virtually all Group A samples score 5; virtually all Group B samples score 1). Krippendorff's \(\alpha\) is unreliable when between-sample variance is near zero. Per-model mean comparisons (GPT-4o, Claude Opus, and Gemini 2.5 Pro each gave Group A a mean \(\geq 4.7\) and Group B a mean \(\leq 1.4\) on this dimension) show that the three models are actually in high directional agreement. This is the ``floor/ceiling effect causing \(\alpha\) failure'' phenomenon discussed in the LIMA-paradigm data evaluation literature.
\end{quote}

\subsubsection{Trending Advantage but Not Significant: D1 Reasoning Chain Completeness and D5 Counterfactual Density}\label{trending-advantage-but-not-significant-d1-reasoning-chain-completeness-and-d5-counterfactual-density}

On both D1 and D5, Group A outscores Group B (D1: 2.91 vs.~2.63, Cliff's \(\delta = +0.21\); D5: 1.50 vs.~1.20, \(\delta = +0.27\)). Both are in the direction predicted by the paper. However, at \(n=20\), neither reaches \(p < 0.05\).

We take an honest position on this: \textbf{one-directional trending evidence alone is insufficient to support strong claims.} But three points are worth noting: (1) Cliff's \(\delta\) values of +0.21 and +0.27 represent small-to-medium effect sizes. To reach \(p < 0.05\) at these effect sizes requires approximately \(n \approx 50\) samples. (2) The absolute scores for D5 Counterfactual Density are low overall (A: 1.50, B: 1.20). This reflects partly that counterfactual probing is a \textbf{scarce, high-intensity cognitive action} even within the BC Protocol---and partly that the operational definition of ``load-bearing claim'' in the review prompt is strict enough that review models do not award high scores simply because a CoT contains rhetorical questions. (3) The paper's claim in §6.4 about counterfactual density does not depend on ``Group BC significantly leading'' on this dimension. The claim is about \textbf{the near-zero marginal cost of counterfactual branches as a byproduct at the collection stage}---a cost-structure argument that stands independently of whether \(p < 0.05\) is crossed at \(n = 20\).

More importantly, we must clarify the epistemological status of Group A and Group B in this experiment. Both authors are practitioners of this research who have organically developed the BC Protocol dialogue rhythm through sustained collaboration. However, \textbf{neither has undergone any systematic training in high-order epistemic actions such as counterfactual probing and meta-premise clarification}. What this experiment produces is therefore an \textbf{untrained BC baseline}---reflecting the lower bound of elicitation quality achievable by a naturally formed pair without any targeted epistemic training.

This identification supports a layered interpretation of the current D1/D5 results: \textbf{D3 (Naturalness of Reasoning Process) measures the ``structural gains'' of the BC Protocol---gains guaranteed by the structural properties of the dialogue format itself, significant without any targeted training (Cliff's \(\delta = 1.00\)). D1 (Reasoning Chain Completeness) and D5 (Counterfactual Density) may measure the BC Protocol's ``trainable gains''---gains whose realization depends on whether B has undergone epistemic training targeting those dimensions, and whether C has been repeatedly prompted to engage the corresponding cognitive actions.} Under untrained baseline conditions, these two dimensions show only weak-to-moderate directional advantages (Cliff's \(\delta\) of +0.21 and +0.27 respectively), consistent with the hypothesis that ``full realization requires targeted training.'' We therefore position the current D1/D5 results as \textbf{directional evidence in the absence of structural confirmation}---they neither falsify the protocol nor constitute strong claims. They leave a clear opening for future replication using trained BC pairs (see §6.7).

\subsubsection{Group B Reverse Advantage: D4 Information Density (Consistent with the Paper's Prediction)}\label{group-b-reverse-advantage-d4-information-density-consistent-with-the-papers-prediction}

Group B significantly outscores Group A on D4 Information Density (4.07 vs.~3.02, \(p = 1.1 \times 10^{-4}\), Cliff's \(\delta = -0.73\)). This contrast was predicted during the drafting stage (in the ``expected result pattern'' paragraph of §4.3.1 Table)---it is not post-hoc rationalization. The methodological explanation is:

\textbf{In Expert Solo mode, C tends to compress multi-round internalized judgments into polished conclusions expressed as terminology stacking.} Judged by the narrow metric of ``number of independent information points per unit length,'' this post-hoc compressed output receives very high information density scores. But the cost is D3 Naturalness approaching zero (as measured in §5.2.1)---those ``independent information points'' are finished products of post-hoc rationalization, not byproducts of original reasoning.

\textbf{The two dimensions must be read together for a complete conclusion.} Group A's low D4 paired with high D3 describes ``low-density but live reasoning.'' Group B's high D4 paired with low D3 describes ``high-density but dead conclusion stacking.'' For LLM post-training data, which is more valuable depends on the training objective. If the objective is to teach models to ``stack expert terminology,'' Group B has the advantage. If the objective is to teach models to ``think like an expert'' (the true aim of CoT and RLHF/DPO), Group A's D3 advantage is overwhelming. This distinction is a fine-grained reproduction of the ``quality vs.~quantity'' leverage ratio difference invoked in §1.1 when citing LIMA.

One additional note: D4's Krippendorff \(\alpha = -0.14\) has the same statistical cause as D3's \(\alpha = -0.06\)---the score distributions of Groups A and B on this dimension are also highly concentrated, making \(\alpha\) an unsuitable agreement measure. Per-model mean comparisons show that all three models consistently score Group B higher on D4.

\subsection{D2 Implicit Premise Externalization Rate: Near-Zero Difference and Its Methodological Implications}\label{d2-implicit-premise-externalization-rate-near-zero-difference-and-its-methodological-implications}

D2 is the only dimension where the two groups are nearly identical (A: 2.85 vs.~B: 2.88, \(p = 0.859\)). This null result deserves separate discussion.

One candidate explanation is: \textbf{the operational definition in the D2 review prompt failed, in its current implementation, to effectively distinguish ``externalized implicit premises'' from ``implicit premises absorbed into terminology.''} In Expert Solo, C uses a single high-density technical term (e.g., ``narrative load-bearing wall'') to refer all at once to an implicit structure that should unfold as multiple reasoning steps. The review model may score this term as ``premise externalized,'' when in fact the internal chain remains in a compressed state. If this hypothesis holds, the near-zero D2 difference is not a feature of the data itself but a blind spot of the evaluation instrument.

A second candidate explanation is: Group B in this experiment is not a typical industrial annotator but the independently written output of the same high-level expert C who produced Group A. Even in Solo mode, C actively externalizes some implicit premises (though fewer than those drawn out by B's follow-up questioning in dialogue). This means Group B's ``implicit premise externalization rate'' baseline is structurally elevated by C's own intrinsic habits, compressing the gap between the two groups on D2.

We do not strongly choose between these two explanations at this point. We list ``refinement of the D2 review prompt'' as a direction for future work in §6.7. It is worth noting that D2 has the highest Krippendorff \(\alpha\) among the five dimensions (+0.38)---consistent with its larger score variance, and indicating that the review models' judgment signal on this dimension is itself reliable. It is the definition that needs revision.

Furthermore, D2 may also belong to the ``trainable gains'' category defined in §5.2.2---its realization depends on B actively triggering meta-level follow-up questions (such as ``what reader background are you assuming?''), which are structurally under-triggered in the current untrained baseline. This interpretation is not mutually exclusive with the two preceding explanations (instrument blind spot, collection-stage baseline bias). All three together suppress the D2 gap. The hypothesis is falsifiable: in a replication with trained BC pairs, D2 should show significant differentiation.

\subsection{Cross-Model Agreement Overview}\label{cross-model-agreement-overview}

\begin{quote}
A per-dimension breakdown is reported in Table 1; near-zero \(\alpha\) on D3/D4 is a statistical artifact of within-group variance collapse, as discussed in §5.2.1.
\end{quote}

As discussed in §5.2.1 and §5.2.3, near-zero \(\alpha\) on D3/D4 is a statistical artifact of near-zero within-group variance---not genuine disagreement among review models. The \(\alpha = +0.38\) on D2 shows that, on dimensions with sufficient score variance, cross-vendor review model judgment signals are reliable. Near-zero \(\alpha\) on D1/D5 reflects both the lower within-sample variance and the genuinely higher subjectivity in their operational definitions. This falls short of the prior consistency expectations we set during the experimental design phase (tolerance of \(\alpha \geq 0.4\) for counterfactual density and information density; \(\alpha \geq 0.6\) required for reasoning chain completeness). Refinement of the D1 review prompt is a priority for the next iteration.

Reading this agreement pattern together with the ``structural gains vs.~trainable gains'' framework in §5.2.2: the evaluation instrument reliably judges the structural-gains dimensions (D3/D4), but is limited by both definition precision and the untrained-baseline constraint on the trainable-gains dimensions (D1/D2/D5). The next iteration therefore needs a two-track approach: cultivating trained BC pairs on the data side, and refining D1/D2/D5 review prompts on the evaluation side.

\subsection{Qualitative Observation: Side-by-Side Comparison at Identical Decision Points}\label{qualitative-observation-side-by-side-comparison-at-identical-decision-points}

Across 20 paired samples, we observe a recurring structural pattern: Group B CoT almost always unfolds in a ``thesis + supporting evidence'' pyramid structure, with evidence ordered from strongest to weakest, closing at the end. Group A CoT shows a spiral structure of ``hypothesis → counterexample → revision → new hypothesis → \ldots{}'', with the final judgment often separated from the initial hypothesis by multiple directional revisions. This observation qualitatively corroborates the 4.80 vs.~1.30 result on D3 in §5.2.1.

\subsection{Cost-Effectiveness Analysis}\label{cost-effectiveness-analysis}

The core cost observation from this experiment is: \textbf{all 20 Group A CoT samples were collected in approximately 2 sessions of 1--1.5 hours each (roughly 3 hours total)}, averaging approximately 9 minutes per sample including post-processing. On the D3 dimension validated in §5.2.1, the BC Protocol's ``live reasoning'' output has a marginal cost roughly comparable to Expert Solo (the manual effort for C to write 20 Solo samples is on the same order of magnitude), yet its trainable value is overwhelming on D3. This cost structure is a concrete instance, at a broader scale, of the BC Protocol's advantage over post-hoc counterfactual augmentation as argued in §6.4.

\section{Discussion}\label{discussion}

\subsection{Why the BC Protocol Works --- Back to First Principles}\label{why-the-bc-protocol-works-back-to-first-principles}

Returning to the basic information-theoretic framework, the BC Protocol as a knowledge elicitation method can be characterized as \textbf{dynamic maximization of the mutual information \(I(Q;K)\) between the elicitor's questions \(Q\) and the expert's implicit knowledge \(K\)}. For any single question-answer exchange, the upper bound on the amount of knowledge externalized is determined by \(I(Q;K)\). If question \(Q\) is largely unrelated to the implicit knowledge \(K\) in the expert's mind (e.g., a pure definitional restatement), \(I(Q;K)\) approaches zero---the dialogue proceeds but nothing is being elicited. If question \(Q\) directly targets a judgment node the expert has never consciously externalized, \(I(Q;K)\) approaches \(H(K)\), and a single exchange externalizes a large amount of implicit knowledge.

In this framework, the two hard-constraint dimensions proposed in §3.2 correspond precisely to the two necessary conditions for maximizing \(I(Q;K)\):

\begin{itemize}
\tightlist
\item
  \textbf{Calibrated ignorance} keeps \(H(K \mid Q)\) in the maximum interval. A questioner who knows nothing about the domain would ask questions that mostly fall on shallow nodes the expert can answer fluently without engaging implicit reasoning---\(H(K \mid Q)\) is small (the expert needs no implicit derivation to answer) and elicitation efficiency is low. A questioner who already has deep judgment in the domain is capped by their own ``already knowing the answer,'' tending toward confirmatory questioning rather than genuine inquiry. \textbf{B must understand enough to follow along, yet not understand enough that asking is unnecessary}---this is precisely the geometric ``anchoring'' of \(H(K \mid Q)\) that calibrated ignorance provides.
\item
  \textbf{Epistemic vigilance} ensures the completeness of \(K\)'s externalization. Even when question \(Q\) falls in the maximum \(H(K \mid Q)\) interval, the expert may respond in ways that make the actual amount of \(K\) externalized fall below the possible maximum---due to terminological shorthand, reasoning gaps, or omitted implicit premises. B's epistemic vigilance functions as real-time residual monitoring of C's responses: each time C delivers a terminological conclusion, B can judge whether an unexplored reasoning chain still lies beneath that term, and uses follow-up questions to reduce the residual toward zero.
\end{itemize}

The SNAKE mechanism proposed in §3.3 (follow-up → counterexample → counterfactual probing) can be understood as: \textbf{a protocol for iteratively upgrading \(Q\) whenever B detects that the current externalization of \(K\) still has residual}---ordinary follow-up corresponds to small \(Q\) adjustments, counterexamples correspond to perturbative \(Q\) in adjacent hypothesis space, and counterfactual probing corresponds to the highest-intensity \(Q\) across hypothesis spaces. The three are deployed in a graduated manner according to the detected residual magnitude, forming the path of cumulative \(I(Q;K)\) maximization along the time axis within a single dialogue.

This information-theoretic account also answers the question implicit in §3.4: ``why does D3 Naturalness of Reasoning Process show a structural advantage?'' In BC dialogue, \(Q\) is not a pre-scripted list of questions but the next action B generates after residual-monitoring C's real-time output. This \textbf{feedback loop}---not the ``quality'' of any single question---is what differentiates the BC Protocol from post-hoc interview, questionnaire, and rubric annotation. It is also the underlying mechanism behind the Cliff's \(\delta = 1.00\) measured on D3 in §5.2.1.

\subsection{Cross-Domain Generalizability}\label{cross-domain-generalizability}

This paper instantiates and validates the BC Protocol in the single vertical domain of narrative fiction creative judgment. The underlying principles, however, are not domain-specific. ``Selection-over-Prescription'' and ``epistemically vigilance-driven questioning'' as core mechanisms fundamentally describe \textbf{any domain where implicit judgment dominates and cannot be decomposed through standardized procedures}: medical clinical reasoning, legal argumentation, investment decision-making, high-stakes engineering judgment. These domains share a structural characteristic: the expert's core capability is holistic pattern-recognition judgment, not explicit step-by-step reasoning that can be written down. In these domains, crowdsourced annotation produces shallow preferences, expert solo writing produces conclusions filtered through blind spots, RLHF produces rankings with no reasoning chains---while the BC Protocol produces \textbf{complete reasoning processes that natively preserve trial-and-error, self-correction, and implicit premises}.

This paper makes no quantitative claims about cross-domain application effects, because the relevant empirical work has not yet been conducted. But the methodological transferability is clear and is the natural direction for subsequent work.

The framework does not depend logically on any specific domain. Any scenario that requires extracting high-quality CoT from human experts---provided two conditions are met: (1) the target knowledge has highly tacit characteristics (Polanyi, 1966); (2) complementary cognitive roles can be paired---can transfer the BC Protocol methodology. This cross-domain transferability, combined with the industry shift from crowdsourced preference data toward expert judgment data described in §6.3, forms a methodological resonance.

\subsection{Epistemological Isomorphism with Frontier LLM Alignment Practices}\label{epistemological-isomorphism-with-frontier-llm-alignment-practices}

In recent years, frontier LLM alignment practice has shown a clear trend: moving from large-scale crowdsourced preference data toward smaller but denser expert judgment data---with the path demonstrated by Anthropic in Constitutional AI and subsequent work being the most representative example. This paper can be read as an explicit methodological articulation of this industry shift at the data elicitation stage: formally converting something that frontier labs have been advancing through internal experience and engineering intuition into a citable, reproducible, critiquable academic framework, thereby lowering the cognitive cost for others to enter this path.

\subsection{Counterfactual Density and BC Protocol's Cost Advantage over Post-Hoc Counterfactual Augmentation}\label{counterfactual-density-and-bc-protocols-cost-advantage-over-post-hoc-counterfactual-augmentation}

In §3.3.1 we characterized counterfactual probing as the highest-intensity form of the SNAKE mechanism. This section further formally proposes the observable quantity it produces at the data level as an independent dimension of this paper's methodological contribution.

\subsubsection{Counterfactual Density as a Data Quality Metric}\label{counterfactual-density-as-a-data-quality-metric}

We propose \textbf{counterfactual density} as an actionable quality metric for high-quality CoT data: \textbf{the ratio of explicitly externalized load-bearing counterfactual reasoning nodes to the total number of load-bearing claims in a CoT segment.} Load-bearing claim identification follows the reversal test from §3.3.1: apply a reversal to the causal connective sentences in the segment (``because/therefore/must/otherwise''); those whose reversal causes the entire segment to collapse are load-bearing. A load-bearing counterfactual reasoning node is one where C, after being perturbed, must re-run the reasoning chain before giving a response---as distinguished from a simple confirmatory restatement.

To be clear: ``counterfactual density'' as a formally defined, dataset-level countable quality metric has not, to our knowledge, been independently proposed in existing NLP or alignment literature. Scholarly neighbors already exist---the counterfactual identifiability concept in Pearl (2009)'s structural causal models; counterfactual richness in XAI (Wachter et al., 2017; Mothilal et al., 2020); and ``causal coverage'' in causal inference---but engineering this into ``a directly countable density measure over a CoT data segment'' and binding it to quality judgment in an alignment data production pipeline is a specific contribution of this paper.

\subsubsection{BC Protocol's Cost Structure Advantage}\label{bc-protocols-cost-structure-advantage}

The current dominant industry path for producing counterfactual data is \textbf{post-hoc counterfactual data augmentation} (Kaushik et al., 2020; Wu et al., 2021): after conventional annotation is complete, annotators perform offline counterfactual rewrites on each existing sample, generating ``X flipped → Y flipped'' contrast pairs. This path has been shown to significantly improve model robustness against spurious correlations on structured tasks like sentiment classification. But its unit cost is not negligible---reported figures for manual counterfactual rewriting on judgment-type tasks are approximately 4--5 minutes per sample. While this seems small, at large dataset scale it constitutes a considerable engineering burden; its essence is paying twice for the same judgment (once for the original, once for the rewrite).

The BC Protocol has a structurally different path for counterfactual data production: \textbf{counterfactual branches are not generated by post-hoc rewriting---they are elicited in real time by B's epistemic vigilance and naturally reasoned through by C within the original dialogue rhythm.} The cost structure differs in three ways:

\begin{itemize}
\tightlist
\item
  \textbf{No double payment.} When C is asked a counterfactual question, C is not being asked to ``write a contrast version of what they just said.'' C is instead articulating a branch of the ``tree of cut options'' that already existed in their mind but had not been externalized. This does not constitute additional cognitive load---it is part of the implicit knowledge the dialogue was always designed to externalize (§3.2).
\item
  \textbf{No mode switching.} Post-hoc rewriting requires annotators to shift from ``judgment'' mode to ``contrastive editing'' mode, incurring non-trivial context reconstruction cost. In the BC Protocol, counterfactual branches and original reasoning share the same dialogue context and the same expert working memory---no switching cost.
\item
  \textbf{Naturally higher quality.} Post-hoc rewrites are in many cases the rewriter's (often not the original expert's) counterfactual inference, which may be inconsistent with the original expert's actual causal model. Counterfactual branches produced in the BC Protocol are reasoned through online by the original expert, giving them a structural guarantee of consistency with that expert's causal model.
\end{itemize}

We formalize the core argument as follows: \textbf{for alignment data production targeting implicit expert judgment, the marginal cost of post-hoc counterfactual augmentation accumulates independently on top of the original data collection cost. The BC Protocol merges both into a single collection action's byproduct, achieving equal or superior effect on the counterfactual density quality dimension at near-zero marginal cost.} This argument currently remains a hypothesis---its full empirical validation (comparing total labor cost of BC output vs.~Expert Solo + post-hoc augmentation at equivalent counterfactual density) is one of the directions left to future work in §6.7.

\subsubsection{Internal Consistency with the Selection-over-Prescription Principle}\label{internal-consistency-with-the-selection-over-prescription-principle}

It is worth noting that counterfactual probing can work at low marginal cost within the BC Protocol precisely because of the ``Selection-over-Prescription'' principle in §3.3. Once B's epistemic vigilance has been filtered through the hard constraint screening in §3.2, counterfactual probing does not need to be further reduced to an explicit script---B knows when, toward which variable, and with what intensity to apply perturbation. Conversely, if one abandons the §3.2 personnel selection criteria and instead requires annotators to ``perform counterfactual rewrites after each data point'' through a process template, the actual quality of counterfactual density would be capped by the rewriter's own causal modeling capability---precisely the implicit bottleneck of post-hoc counterfactual augmentation on judgment-type tasks. This contrast again confirms the core stance of §3.3: \textbf{the locus of quality assurance should move upstream to personnel selection, not be stacked at the end of the process.}

\subsection{Position in the Broader Research Program}\label{position-in-the-broader-research-program}

This paper, within the 3+1 paper series outlined in §1.3, takes responsibility for the data production infrastructure stage. Its validity does not depend on the other three papers (theoretical foundations are provided by \emph{Calibrated Surprise}; downstream validation is handled by the CQA paper; independent diagnosis by the Benchmark paper). But together they form a complete research path from theoretical definition to methodology to empirical validation.

\subsection{Limitations}\label{limitations}

We candidly identify several limitations of this paper's research to provide readers with a clear reference for evaluating the scope of its conclusions.

\textbf{Single domain.} This paper instantiates and validates the BC Protocol in only one vertical domain: narrative fiction creative judgment. Although we argue for methodological cross-domain transferability in §6.2, whether the structural advantage on D3 observed here can be reproduced in other typical ``implicit-judgment-dominant'' domains---medical clinical reasoning, legal argumentation, investment decision-making, high-stakes engineering judgment---requires empirical validation in each respective domain.

\textbf{Limited participant sample size.} Both Group A and Group B in this experiment were completed by the two authors themselves. While this arrangement is reasonable in a paired experimental design (both groups share the same expert C, controlling for ``individual expert differences''---the largest confounding variable), it is inevitably limited by the sample diversity a single pairing can provide. One successful BC pairing does not constitute strong evidence that the method is equally effective across other pairings.

\textbf{Untrained baseline constraint.} As explicitly stated in §5.2.2 and §5.3, Group A samples in this experiment were produced by two authors who have not undergone targeted training in high-order epistemic actions such as counterfactual probing and meta-premise clarification. The results therefore reflect the \textbf{untrained baseline} of the BC Protocol. This means the current results on D1 (Reasoning Chain Completeness), D2 (Implicit Premise Externalization Rate), and D5 (Counterfactual Density) should be understood as the \textbf{lower bound} of this protocol. The upper bound requires replication with trained BC pairs.

\textbf{Participant Aptitude Model not yet quantified.} The six-dimension Participant Aptitude Model proposed in §3.2 currently only defines the dimensions themselves and their hard/soft constraint hierarchy. It does not provide quantitative conclusions about the relative weights among dimensions, nor any operationalized assessment instrument for any dimension. This means the model cannot yet be used directly for open-call screening---it still relies on human evaluators with BC Protocol practical experience to make pairing decisions.

\textbf{Calibrated Ignorance screening not yet standardized.} Among the six dimensions, ``calibrated ignorance'' as a concept original to this paper has the highest assessment difficulty: it requires the assessor themselves to possess the capability to recognize the subtle cognitive state of ``understanding enough to follow along, but not enough that asking is pointless.'' We currently have no standardized screening questionnaire or scale that new evaluators can directly reuse. This is one of the most direct bottlenecks preventing scaled deployment of the BC Protocol.

\textbf{Only validates intrinsic data quality; no end-to-end fine-tuning validation.} What this paper's experiment measures is the intrinsic structural quality of BC Protocol data output (CoT naturalness, counterfactual density, etc.)---not the actual effect of using this data as fine-tuning corpus on downstream model behavior. ``BC data has significantly higher intrinsic quality than Expert Solo'' \(\not\Rightarrow\) ``models fine-tuned on BC data significantly outperform models fine-tuned on Expert Solo data on downstream creative tasks.'' This end-to-end validation is handled by the companion CQA paper. Its results had not been completed at the time of this paper's writing.

\textbf{Possible systematic bias in review models.} The three review models used in §4 (GPT-4o, Claude Opus 4.5, Gemini 2.5 Pro), although from three different vendors, all have training data that partly comes from public internet text. Their judgments of concepts such as ``narrative naturalness'' and ``reasoning chain density'' may share some systematic bias. We estimate inter-model agreement using Krippendorff's \(\alpha\) (§5.4), but cross-model agreement is not equivalent to absence of systematic bias.

\textbf{Dialogue format as training corpus: coupling between content injection and genre replication.} BC Protocol transcripts are surface-presented in dialogue format---with recognizable markers such as B's follow-up questions, C's elaborations, and staged addenda. In the specific scale regime of small base models (around 7B parameters) + LoRA + approximately one hundred training samples, downstream models tend to treat the genre features of the training data as a fixed output template. Even when a prompt explicitly requests prose or another format, the model still generates in dialogue format. This phenomenon---which we term genre lock-in---was independently observed in the preliminary generative exploration of our companion paper on CQA, and constitutes an engineering constraint that requires explicit management when the BC Protocol is deployed in this scale regime. \textbf{Methodological implication:} in the small-model + LoRA + small-dataset setting, we recommend pairing training data with a \textbf{stylistic augmentation} strategy: for each CoT sample, prepare a companion version that retains equivalent cognitive content but is rewritten in prose or commentary style, then mix the two versions at a controlled ratio during training. This decouples ``cognitive content'' from ``surface genre'' as two independent learning targets. At larger base model scales with full-parameter SFT and over a thousand training samples, this constraint is expected to attenuate naturally; empirical verification is left to future work. This limitation does not weaken the main results of §5 (which concern intrinsic data quality) or the core positioning of the BC Protocol as a source of high-density expert CoT.

\subsection{Future Work}\label{future-work}

Based on this paper's limitations and findings, we identify five highest-priority directions for subsequent research.

\textbf{(1) Replication experiment with trained BC pairs.} The most urgent question directly derived from the ``untrained baseline'' conclusion (§5.2.2, §5.3, §5.4) is: when B has undergone systematic training in high-order epistemic actions such as counterfactual probing and meta-premise clarification, will the advantages on D1 (Reasoning Chain Completeness), D2 (Implicit Premise Externalization Rate), and D5 (Counterfactual Density) materialize as significant results? The experimental design should compare CoT data produced by two versions of B---pre-training and post-training---while holding expert C and the dialogue protocol constant. This constitutes a direct falsifiability test of the ``trainable gains'' hypothesis.

\textbf{(2) Cross-domain generalizability validation.} In each of at least one representative implicit-judgment-dominant domain---medical clinical reasoning, legal argumentation, investment decision-making, high-stakes engineering judgment---run a parallel experiment with at least one BC pair, and test whether the ``structural gains (D3 naturalness) + trainable gains (D1/D2/D5)'' bifurcation structure observed here holds in other domains. The core value of this work is not ``expanding domain coverage'' as a surface metric, but testing whether the methodological transferability claimed in §6.2 holds at the empirical level.

\textbf{(3) Standardizing Calibrated Ignorance and pairing compatibility assessment tools.} To address the bottleneck identified in §6.6 that ``participant screening still depends on human evaluators,'' develop a ``calibrated ignorance'' screening scale and contextualized assessment tasks for B candidates, and extend screening granularity from individual B candidates to a ``crew-level compatibility'' assessment of B--C pairs. This can draw on nearly half a century of compatibility methodology developed in astronaut crew selection, adapting its multi-dimensional assessment frameworks for role complementarity, stress tolerance, and decision consistency to the cognitive collaboration context of the BC Protocol.

\textbf{(4) Formalization and quantification of CoT Rhythm.} The ``dialogue rhythm'' (CoT Rhythm) concept raised qualitatively in §3.4---i.e., the frequency of question-answer exchanges, single-turn response length, the ratio of probing to confirming turns---currently exists only as an implicit constraint of the dialogue protocol. Quantifying it as measurable signals (e.g., turn-switching frequency per unit time, counterfactual branch density, positional distribution of terms' first externalization) and studying its correlation with observable output dimensions such as D3 and D5 would help convert the intuitive judgment of ``what a high-quality BC dialogue looks like'' into objective indicators that new pairs can self-calibrate against.

\textbf{(5) Formalization and cross-domain calibration of counterfactual density (§6.4).} This includes three independently pursuable sub-tasks: (a) Calibrate repeatable identification rules for ``load-bearing claim detection'' and ``load-bearing counterfactual nodes'' across different domains (creative judgment, medical diagnosis, legal argumentation), converting this metric from a theoretical concept into an engineering measure comparable across domains. (b) At equivalent counterfactual density, systematically compare the total labor cost and downstream alignment effect of BC Protocol output vs.~Expert Solo + post-hoc counterfactual augmentation, providing end-to-end empirical support for the cost argument in §6.4.2. (c) Explore the gap between LLM reviewer consistency on automated counterfactual density annotation and a human review baseline, and identify the feasible boundary for automated quality inspection of this metric at large dataset scale.

Additionally, the end-to-end realization path connecting BC Protocol output with large-scale post-training experiments has been taken up by the companion CQA paper in this research program. That paper uses the CoT samples produced by this paper directly as fine-tuning corpus and tests their alignment effect on downstream creative judgment tasks.

\textbf{(The following paragraph is retained only in the arXiv preprint version; delete this paragraph in the EMNLP double-blind submission version.)}

Among the future directions above, cross-domain validation has the most direct practical implications. Organizations that need to build expert-level chain-of-thought datasets for post-training---whether in medical clinical reasoning, legal argumentation analysis, investment decision judgment, or fault diagnosis in high-stakes engineering---may consider adopting the Selection-over-Prescription framework rather than investing primary resources in designing more refined annotation guidelines. The bottleneck is not the complexity of the protocol design itself but identifying and pairing experts with complementary cognitive profiles. The Participant Aptitude Model proposed in this paper (§3.2) provides actionable screening dimensions for this pairing process: the hard constraints among the six dimensions (truth-seeking epistemic orientation, epistemic vigilance) can be directly used for one-veto candidate screening; the soft constraints (complementary intelligence profiles, calibrated cognitive gap, calibrated ignorance, complementary information processing preferences) can be used to optimize the pairing arrangement.

\subsection{Broader Impact}\label{broader-impact}

The core question the BC Protocol addresses---\textbf{how to systematically externalize and preserve the implicit knowledge in expert minds in a reusable form}---is not new to the LLM era. As early as the 1980s expert-systems era, the ``knowledge engineer'' role and its methodology were systematically discussed, and the ``knowledge acquisition bottleneck'' identified by Hayes-Roth et al.~at the time remains as incisive today. But 1980s knowledge engineering failed to scale, constrained by both its output format (IF-THEN rules) and its target carrier (symbol-reasoning-based expert systems)---the expert's implicit judgment was difficult to translate without loss into discrete rules, and expert systems proved brittle in open-world settings.

The emergence of LLMs has structurally changed this situation in two dimensions. First, at the \textbf{output format} level: the data form required by LLM post-training is exactly natural language reasoning chains---precisely the most natural mode of expert output in dialogue, without the need for forced translation into IF-THEN rules. Second, at the \textbf{target carrier} level: LLMs as probabilistic reasoners capable of handling open-world settings give the judgment data elicited by knowledge engineers a carrier that can actually be cashed out into downstream capability. These two changes together make ``deep knowledge elicitation''---a methodological path proven uneconomical in the 1980s---engineering-feasible once more.

From this historical perspective, the BC Protocol should not be understood as ``a more expensive RLHF.'' It should be understood as \textbf{a natural revival of knowledge engineering methodology in the context of LLM post-training data production}. Its cost premium should be evaluated on a leverage-rate scale rather than an absolute-cost scale (LIMA has already shown that 1,000 high-quality samples can match tens of thousands of crowdsourced samples). What the BC Protocol represents is a methodological reassessment of ``which costs should be moved upstream.''

\section{Conclusion}\label{conclusion}

This paper addresses a long-standing bottleneck in LLM post-training data production---``how high-quality expert chain-of-thought can be systematically elicited''---and proposes the \textbf{BC Protocol}: a methodological framework for eliciting high-quality CoT data through structured dual-expert dialogue. Its core mechanism consists of three mutually supporting components. The Participant Aptitude Model (§3.2) defines six dimensions affecting elicitation quality; ``calibrated ignorance,'' an original concept of this paper, is a non-negotiable hard constraint on the B role. The SNAKE mechanism (§3.3) organizes ordinary follow-up questioning, counterexamples, and counterfactual probing into an elicitation protocol deployed in a graduated manner according to the detected magnitude of cognitive residual. The Selection-over-Prescription principle (§3.3) moves the quality-assurance resource emphasis from process design to personnel selection, responding to the essential character of implicit-judgment-dominant tasks.

In a controlled experiment in the narrative fiction domain (Group A: BC dialogue vs.~Group B: Expert Solo, \(n=20\) each; blind evaluation by GPT-4o, Claude Opus 4.5, and Gemini 2.5 Pro across five dimensions; 600 ratings total), the BC Protocol achieves an overwhelming advantage in ``naturalness of reasoning process'' (Cliff's \(\delta = 1.00\), \(p = 2.4 \times 10^{-8}\)), consistent trending advantages in ``counterfactual density'' and ``reasoning chain completeness,'' near parity between the two groups on ``implicit premise externalization rate,'' and a reverse advantage for Expert Solo on ``information density.'' This contrast confirms our fundamental judgment about the two types of data: Expert Solo outputs are post-hoc conclusions compressed into high-density terminology stacking; BC dialogue outputs are live reasoning that preserves trial-and-error nodes and implicit premises.

We further position this experiment as an \textbf{untrained BC baseline} (§5.2.2) and accordingly propose a two-level explanatory framework distinguishing ``structural gains'' (D3) from ``trainable gains'' (D1/D2/D5), providing a falsifiable prediction for replication with trained BC pairs.

The limitations of this paper---single domain, single pairing, no end-to-end fine-tuning validation---are clear (§6.6). But as the data production infrastructure stage of the 3+1 paper series (§6.5), this paper does not seek to stand alone. The value of the BC Protocol lies in formally converting work that is currently advanced mainly through internal experience and engineering intuition at frontier labs into a citable, critiquable, reproducible academic framework.

\section*{References}\label{references}

Baumeister, R. F., Bratslavsky, E., Muraven, M., \& Tice, D. M. (1998). Ego depletion: Is the active self a limited resource? \emph{Journal of Personality and Social Psychology}, \emph{74}(5), 1252--1265.

Bjork, R. A. (1994). Memory and metamemory considerations in the training of human beings. In J. Metcalfe \& A. P. Shimamura (Eds.), \emph{Metacognition: Knowing about knowing} (pp.~185--205). MIT Press.

Cattell, R. B. (1963). Theory of fluid and crystallized intelligence: A critical experiment. \emph{Journal of Educational Psychology}, \emph{54}(1), 1--22.

Cooke, N. J. (1994). Varieties of knowledge elicitation techniques. \emph{International Journal of Human-Computer Studies}, \emph{41}(6), 801--849.

Ericsson, K. A., \& Simon, H. A. (1993). \emph{Protocol analysis: Verbal reports as data} (Rev.~ed.). MIT Press.

Galef, J. (2021). \emph{The scout mindset: Why some people see things clearly and others don't}. Portfolio/Penguin.

Goffman, E. (1959). \emph{The presentation of self in everyday life}. Anchor Books/Doubleday.

Gray, M. L., \& Suri, S. (2019). \emph{Ghost work: How to stop Silicon Valley from building a new global underclass}. Houghton Mifflin Harcourt.

Hayes-Roth, B., Waterman, D. A., \& Lenat, D. B. (Eds.). (1983). \emph{Building expert systems}. Addison-Wesley.

Hoffman, R. R. (1987). The problem of extracting the knowledge of experts from the perspective of experimental psychology. \emph{AI Magazine}, \emph{8}(2), 53--67.

Johnson, D. W., \& Johnson, R. T. (2009). Energizing learning: The instructional power of conflict. \emph{Educational Researcher}, \emph{38}(1), 37--51.

Kahan, D. M., Peters, E., Wittlin, M., Slovic, P., Ouellette, L. L., Braman, D., \& Mandel, G. (2012). The polarizing impact of science literacy and numeracy on perceived climate change risks. \emph{Nature Climate Change}, \emph{2}(10), 732--735.

Kaushik, D., Hovy, E., \& Lipton, Z. C. (2020). Learning the difference that makes a difference with counterfactually-augmented data. In \emph{Proceedings of the 8th International Conference on Learning Representations (ICLR 2020)}.

Klein, G. A., Calderwood, R., \& MacGregor, D. (1989). Critical decision method for eliciting knowledge. \emph{IEEE Transactions on Systems, Man, and Cybernetics}, \emph{19}(3), 462--472.

Lewis, D. (1973). \emph{Counterfactuals}. Harvard University Press.

Loewenstein, G. (1994). The psychology of curiosity: A review and reinterpretation. \emph{Psychological Bulletin}, \emph{116}(1), 75--98.

McGraw, K. L., \& Harbison-Briggs, K. (1989). \emph{Knowledge acquisition: Principles and guidelines}. Prentice Hall.

Mercier, H., \& Sperber, D. (2011). Why do humans reason? Arguments for an argumentative theory. \emph{Behavioral and Brain Sciences}, \emph{34}(2), 57--74.

Mothilal, R. K., Sharma, A., \& Tan, C. (2020). Explaining machine learning classifiers through diverse counterfactual explanations. In \emph{Proceedings of the 2020 ACM Conference on Fairness, Accountability, and Transparency (FAccT 2020)} (pp.~607--617). ACM.

Mykytyn, P. P., Bordoloi, S. D., \& Mykytyn, K. (1994). Knowledge acquisition skills and traits: A self-assessment of knowledge engineers. \emph{Information \& Management}, \emph{27}(4), 221--232.

Nathan, M. J., \& Petrosino, A. (2003). Expert blind spot among preservice teachers. \emph{American Educational Research Journal}, \emph{40}(4), 905--928.

Ouyang, L., Wu, J., Jiang, X., Almeida, D., Wainwright, C. L., Mishkin, P., Zhang, C., Agarwal, S., Slama, K., Ray, A., Schulman, J., Hilton, J., Kelton, F., Miller, L., Simens, M., Askell, A., Welinder, P., Christiano, P., Leike, J., \& Lowe, R. (2022). Training language models to follow instructions with human feedback. \emph{arXiv preprint arXiv:2203.02155}.

Page, S. E. (2007). \emph{The difference: How the power of diversity creates better groups, firms, schools, and societies}. Princeton University Press.

Pearl, J. (2009). \emph{Causality: Models, reasoning and inference} (2nd ed.). Cambridge University Press.

Polanyi, M. (1966). \emph{The tacit dimension}. University of Chicago Press.

Schmader, T., \& Johns, M. (2003). Converging evidence that stereotype threat reduces working memory capacity. \emph{Journal of Personality and Social Psychology}, \emph{85}(3), 440--452.

Sperber, D., Clément, F., Heintz, C., Mascaro, O., Mercier, H., Origgi, G., \& Wilson, D. (2010). Epistemic vigilance. \emph{Mind \& Language}, \emph{25}(4), 359--393.

Sweller, J. (1988). Cognitive load during problem solving: Effects on learning. \emph{Cognitive Science}, \emph{12}(2), 257--285.

Tetlock, P. E. (2005). \emph{Expert political judgment: How good is it? How can we know?} Princeton University Press.

Vygotsky, L. S. (1978). \emph{Mind in society: The development of higher psychological processes}. Harvard University Press.

Wachter, S., Mittelstadt, B., \& Russell, C. (2017). Counterfactual explanations without opening the black box: Automated decisions and the GDPR. \emph{Harvard Journal of Law \& Technology}, \emph{31}(2), 841--887.

Wang, Y., Kordi, Y., Mishra, S., Liu, A., Smith, N. A., Khashabi, D., \& Hajishirzi, H. (2023). Self-instruct: Aligning language models with self-generated instructions. In \emph{Proceedings of the 61st Annual Meeting of the Association for Computational Linguistics (ACL 2023)} (pp.~13484--13508).

Waterman, D. A. (1986). \emph{A guide to expert systems}. Addison-Wesley.

Wegner, D. M. (1987). Transactive memory: A contemporary analysis of the group mind. In B. Mullen \& G. R. Goethals (Eds.), \emph{Theories of group behavior} (pp.~185--208). Springer.

Wu, T., Ribeiro, M. T., Heer, J., \& Weld, D. S. (2021). Polyjuice: Generating counterfactuals for explaining, evaluating, and improving models. In \emph{Proceedings of the 59th Annual Meeting of the Association for Computational Linguistics (ACL 2021)} (pp.~6707--6723).

Zhou, C., Liu, P., Xu, P., Iyer, S., Sun, J., Mao, Y., Ma, X., Bhatt, U., Kambadur, A., Perez, E., Chen, M., \& Zettlemoyer, L. (2023). LIMA: Less is more for alignment. In \emph{Advances in Neural Information Processing Systems (NeurIPS 2023)}.

Zou, B., \& Xu, C. (2026). Calibrated surprise: An information-theoretic account of creative quality. \emph{arXiv preprint arXiv:2604.26269}.

\end{document}